\documentclass[10pt,journal,compsoc]{IEEEtran}

\ifCLASSOPTIONcompsoc
\usepackage[nocompress]{cite}
\else
\usepackage{cite}
\fi

\ifCLASSINFOpdf
\usepackage[pdftex]{graphicx}
\DeclareGraphicsExtensions{.pdf,.jpeg,.png}
\DeclareGraphicsExtensions{.eps}
\usepackage{epstopdf}
\else
\fi

\ifCLASSOPTIONcompsoc
\usepackage[caption=false,font=footnotesize,labelfont=sf,textfont=sf]{subfig}
\else
\usepackage[caption=false,font=footnotesize]{subfig}
\fi
\usepackage{booktabs}
\usepackage{amssymb}

\usepackage{mathrsfs}
\usepackage{epsfig}
\usepackage{graphicx}
\usepackage{amsmath}
\usepackage{cite}
\usepackage{enumerate}
\usepackage{cases}
\usepackage{multirow}
\usepackage{verbatim}
\usepackage{amssymb}
\usepackage{CJK}
\usepackage{algorithmicx}
\usepackage{bbm}
\usepackage{color}
\usepackage{bm}
\usepackage{booktabs}
\usepackage{colortbl}
\usepackage{float}
\usepackage{ragged2e}  % use justify
\usepackage{csquotes}
\usepackage[breaklinks=true,bookmarks=false]{hyperref}
\usepackage{diagbox}
\usepackage{threeparttable}
\usepackage{xspace}

\usepackage{array}
\usepackage[table]{xcolor}
\usepackage{overpic}
\usepackage{textcomp}
\usepackage{contour}
\usepackage[table]{xcolor}
\usepackage{threeparttable}
\usepackage{tablefootnote}

\usepackage{silence}
\makeatletter
\newcommand{\thickhline}{%
	\noalign {\ifnum 0=`}\fi \hrule height 1pt
	\futurelet \reserved@a \@xhline
}
\makeatother

\makeatletter
\DeclareRobustCommand\onedot{\futurelet\@let@token\@onedot}
\def\@onedot{\ifx\@let@token.\else.\null\fi\xspace}

\begin{document}

	\title{Semi-Supervised Coupled Thin-Plate Spline Model for Rotation Correction and Beyond}

	\author{Lang Nie, Chunyu Lin, Kang Liao, Shuaicheng Liu, Yao Zhao,~\IEEEmembership{Fellow,~IEEE}
		
	% \thanks{Corresponding author: Chunyu Lin}
	\thanks{Lang Nie, Chunyu Lin, and Yao Zhao are with the Institute of Information Science, Beijing Jiaotong University, Beijing 100044, China, and also with the Beijing Key Laboratory of Advanced Information Science and Network Technology, Beijing 100044, China (e-mail: nielang@bjtu.edu.cn, cylin@bjtu.edu.cn, yzhao@bjtu.edu.cn).}
    \thanks{Kang Liao is with the School of Computer Science and Engineering, Nanyang Technological University, Singapore (e-mail:  kang.liao@ntu.edu.sg).}
    \thanks{Shuaicheng Liu is with the School of Information and Communication Engineering, University of Electronic Science and Technology of China, Chengdu, 611731, China (e-mail: liushuaicheng@uestc.edu.cn).}
 \thanks{This work was supported by the National Natural Science Foundation of China (NSFC) under Grant 62172032.\\
 (Corresponding author: Chunyu Lin.)}

	}
	% The paper headers
	%\markboth{IEEE TRANSACTIONS ON PATTERN ANALYSIS AND MACHINE INTELLIGENCE}%
	\markboth{}
	{Shell \MakeLowercase{\textit{et al.}}: Bare Demo of IEEEtran.cls for Computer Society Journals}

	\IEEEtitleabstractindextext{%
		\justify  % crucial to justify the abstract
		\begin{abstract}
\label{sec:Abstrat}
Thin-plate spline (TPS) is a principal warp that allows for representing elastic, nonlinear transformation with control point motions. With the increase of control points, the warp becomes increasingly flexible but usually encounters a bottleneck caused by undesired issues, e.g., content distortion. In this paper, we explore generic applications of TPS in single-image-based warping tasks, such as rotation correction, rectangling, and portrait correction. To break this bottleneck, we propose the coupled thin-plate spline model (CoupledTPS), which iteratively couples multiple TPS  with limited control points into a more flexible and powerful transformation. Concretely, we first design an iterative search to predict new control points according to the current latent condition. Then, we present the warping flow as a bridge for the coupling of different TPS transformations, effectively eliminating interpolation errors caused by multiple warps. Besides, in light of the laborious annotation cost, we develop a semi-supervised learning scheme to improve warping quality by exploiting unlabeled data. It is formulated through dual transformation between the searched control points of unlabeled data and its graphic augmentation, yielding an implicit correction consistency constraint. Finally, we collect massive unlabeled data to exhibit the benefit of our semi-supervised scheme in rotation correction. Extensive experiments demonstrate the superiority and universality of CoupledTPS over the existing state-of-the-art (SoTA) solutions for rotation correction and beyond. The code and data are available at \url{https://github.com/nie-lang/CoupledTPS}. 
\end{abstract}

		\begin{IEEEkeywords}
			Thin-plate spline, Semi-supervised learning, Rotation correction, Rectangling, Portrait correction.
	\end{IEEEkeywords}}
	
	% make the title area
	\maketitle
	
	\IEEEdisplaynontitleabstractindextext
	
	\IEEEpeerreviewmaketitle
	
	% !TEX root = ../main.tex

%--------------------------------------------------------
\IEEEraisesectionheading{\section{Introduction}
	\label{sec:Introduction}}
%--------------------------------------------------------
\IEEEPARstart{T}{hin-plate} spline \cite{bookstein1989principal} describes the deformations specified by finitely many point-correspondences in an irregular spacing between a flat image and a warped one. 
Compared with affine (6 DoF) or homography (8 DoF) \cite{hartley2003multiple}, TPS is nonlinear and much more flexible.
It has been widely used in various computer vision tasks, such as image registration \cite{sprengel1996thin}, image stitching \cite{li2017parallax, nie2023parallax}, image animation \cite{zhao2022thin}, and document dewarping \cite{xue2022fourier, jiang2022revisiting, xie2021document}, etc. 

\begin{figure}[!t]
  \centering
  \includegraphics[width=.47\textwidth]{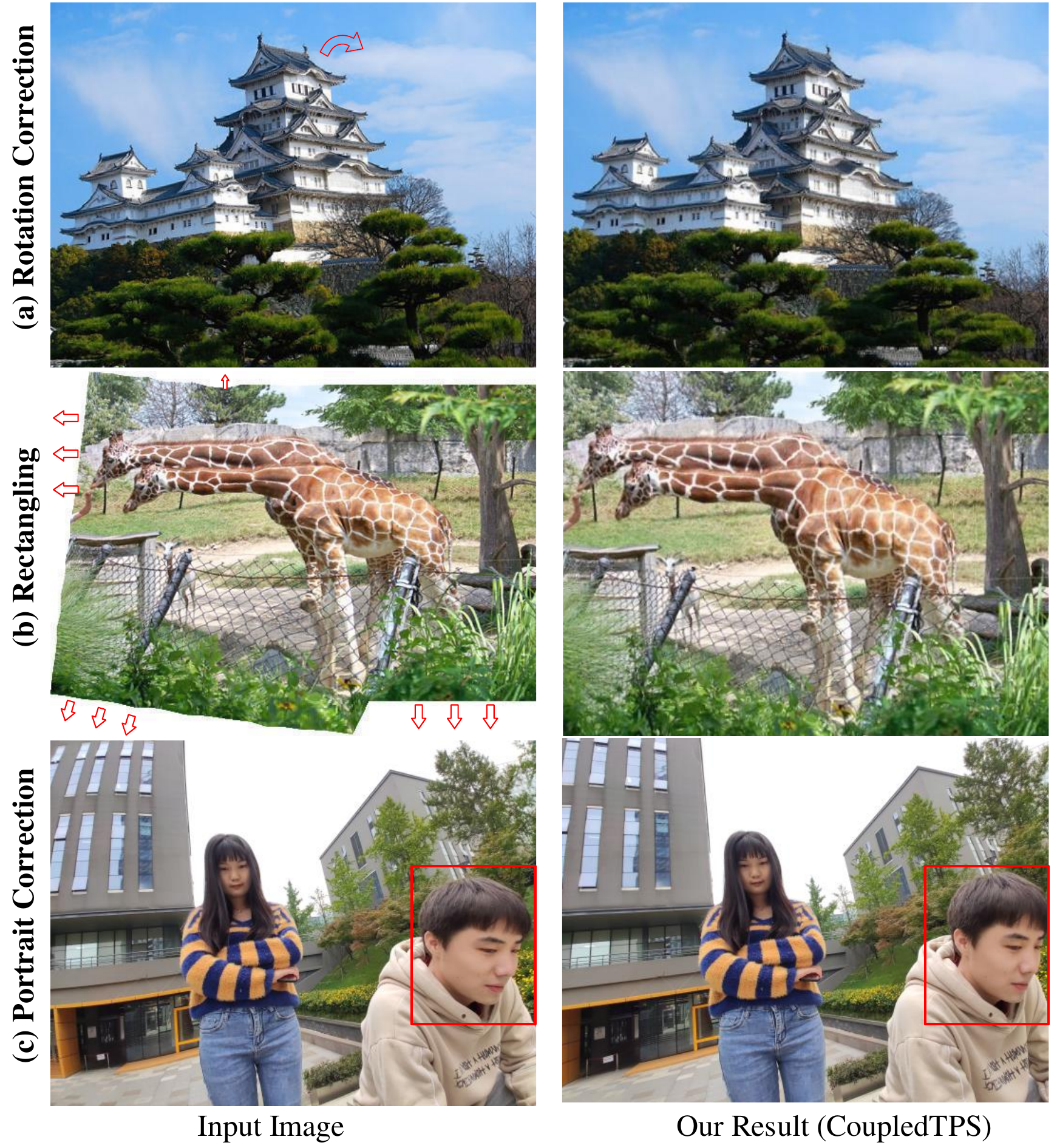}
  %\vspace{-20pt}
  \caption{Three examples of our method. The proposed CoupledTPS corrects the 2D in-plane tilt, irregular boundaries, and wide-angle portrait via a unified warping framework.}
  \label{fig:fig1}
  \vspace{-0.4cm}
\end{figure}

In this paper, we focus on its applications in single-image-based warping tasks, e.g., rotation correction \cite{nie2023deep}, rectangling \cite{nie2022deep}, and wide-angle portrait correction \cite{tan2021practical, zhu2022semi}, as depicted in Fig. \ref{fig:fig1}. Among these tasks, rotation correction aims to rectify the 2D in-plane tilt (roll), rectangling tries to stretch image contents to get rectangular boundaries and portrait correction focuses on corrections of the human faces distorted by camera lenses. Despite the different goals, these tasks are essentially supposed to predict content-aware graphic warps. To this end, we propose to leverage thin-plate spline as a basic transformation for all these tasks. However, as control points increase, the performance usually rises first and is then stuck when the control points increase to a certain quantity. Richer control points endow this transformation with better flexibility, but they also produce more complex and fragile inter-point connections. In particular, when the control points representing the same object (e.g., a straight line) change inconsistently, the image content is prone to local distortion.

% CoupledTPS
To break this performance bottleneck, we propose the coupled thin-plate spline model (CoupledTPS), which aims to further improve flexibility and performance by multiple coupling. Different from searching for sufficient control points for a given input image at one time, CoupledTPS continuously couples multiple basic TPS transformations through multiple iterations, in which each iteration is tailored to search for new control points according to the current condition. To reduce the computational complexity, we define the low-resolution latent condition to replace the high-resolution input image. Concretely, each iteration takes the latent condition as input and we only need to update it in the latent condition space every iteration. To mitigate cumulative interpolation errors arising from multiple warps, we utilize warping flows as a unifying bridge to cohesively couple multiple TPS transformations. In detail, we integrate the previously coupled warp from the previous conditions and the new warp of new predicted control points from the current condition into a more flexible and powerful one. Moreover, each coupling shares network weights, which indicates CoupledTPS does not introduce new parameters in spite of several couplings. 
We evaluate the effectiveness of the proposed CoupledTPS in the tasks of rotation correction, rectangling, and portrait correction. The results show that CoupledTPS not only breaks the performance bottleneck of TPS successfully but also outperforms the existing SoTA methods for these tasks with a unified framework.

%半监督
In addition, we notice all rotation correction solutions \cite{nie2023deep, nie2022deep} adopt the fully supervised learning scheme with a large reliance on laborious annotations.
%There is a fact that all learning-based rotation correction works \cite{nie2023deep, nie2022deep} adopt supervised learning. In light of the laborious annotation cost, 
We propose the first semi-supervised learning strategy to relieve the urgent demand for expensive labels. Given an unlabeled instance, we augment it with rotation-related graphic warping \cite{he2013content}. The semi-supervised learning is developed based on correction consistency: the unlabeled instance and its augmentation have consistent correction results. To this end, the two sets of predicted control points distributed on the unlabeled instance and its augmentation should be one-to-one correspondences. Subsequently, these two instances can establish the dual transformation according to their control points, formulating an implicit correction consistency constraint for unlabeled data. Finally, we collect over 7k unlabeled samples to validate our semi-supervised learning strategy. The results show it not only enhances the warping quality on the labeled data but also assists in generalizing to other domains of unlabeled data.

To sum up, the major contributions center around:
\vspace{-0.2cm}
\begin{itemize}
    \item We propose the coupled thin-plate spline model, termed CoupledTPS, to break the performance bottleneck of TPS. It iteratively integrates multiple basic TPS transformations with limited control points in the latent condition space and couples them into a whole with better flexibility and performance.
    
    \item We develop a semi-supervised learning strategy to relieve the annotation reliance. It formulates an implicit constraint based on correction consistency, promoting both performances on the labeled data and generalization on the unlabeled data.
    
    \item Extensive (over 7k) unlabeled data is collected to validate the benefit of the semi-supervised strategy. Moreover, the proposed CoupledTPS outperforms the SoTA solutions in multiple single-image-based warping tasks, yielding a unified learning framework with high accuracy.
    
 \end{itemize}

% In the following sections, we discuss and analyze various aspects of learning-based camera calibration. The remainder of this paper is organized as follows. In Section~\ref{sec2}, we provide the concrete learning paradigms and learning strategies of the learning-based camera calibration. Subsequently, we introduce and discuss the specific methods based on the standard camera model, distortion model, cross-view model, and cross-sensor model in Section~\ref{sec:pure}, Section~\ref{sec:distortion}, Section~\ref{sec:projection}, and Section~\ref{sec:hybrid}, respectively (see Figure~\ref{fig:taxonomy}). The collected benchmark for calibration methods is depicted in Section~\ref{sec:evaluation}. Finally, we conclude the learning-based camera calibration and suggest the future directions of this community in Section~\ref{sec:Future}.

\vspace{-0.1cm}

	\section{Related Work}
\label{sec:relatedwork}
As our goal is to design a unified learning framework for single-image-based warping tasks, we review the related tasks here, \textit{e.g.}, rotation correction, rectangling, and wide-angle portrait correction. 

\subsection{Rotation Correction}
Rigid rotation operation inclines the whole image with a certain angle. It produces rotated rectangular boundaries, which are no longer suitable for the screen of common display devices (\textit{e.g.}, TVs, mobile phones, etc.). To preserve the regular boundaries, He \textit{et al.} proposed content-aware rotation \cite{he2013content} to rotate the particular contents instead of the whole image. It detected line segments by LSD \cite{von2008lsd}, forced them to rotate to a certain angle, and preserved the contents and boundaries. However, it can only preserve linear structures such as pillars, buildings, etc. When there is a demand to correct a tilted image to the horizontal condition, the specific rotation angle is necessary but unavailable. To acquire this angle prior, whether it is calibrated manually or by automated algorithms \cite{xian2019uprightnet,lee2013automatic, do2020surface}, the whole process becomes cumbersome and prone to accumulate errors.

To simplify this tedious workflow and free from the constriction of angle prior, Nie \textit{et al.} proposed a new and practical computer vision task, named rotation correction \cite{nie2023deep}. 
This task aims to automatically correct the \underline{2D in-plane tilt} (roll) with \underline{high content fidelity} (preserving contents and boundaries) without the angle prior. The corrected results have the perceptually horizontal content and regular rectangular boundaries to fit common display devices.
To this end, Nie \textit{et al.} leveraged a mesh-to-flow transformation network and a rotation correction dataset (DRC-D) with massive triplets of tilted images, tilted angles, and corresponding horizontally corrected labels. 
%Then Nie \textit{et al.} leveraged a mesh-to-flow transformation network to automatically correct the 2D in-plane tilt (roll) in high content fidelity without an angle prior.

\subsection{Rectangling}
Rectangling aims to obtain regular rectangular boundaries from irregular ones (\textit{e.g.}, stitched image\cite{he2013rectangling}, rigidly rotated image\cite{nie2023deep}, and rectified wide-angle image \cite{liao2023recrecnet}) via warping. Usually, it preserves the contents of significant objects and conducts warping operations such as stretching or flattening in other regions (\textit{e.g.}, the lake, sky, etc.), making the distortions visually unnoticeable. 

To preserve these significant objects, traditional algorithms \cite{he2013rectangling, wu2022rectangling, zhang2020content, li2015geodesic} tended to define the straight or geodesic lines as the salient properties and encourage them to keep the original characteristics, \textit{e.g.}, the straight line should keep straight and the parallel lines should keep parallel. However, these hand-craft features cannot deal with non-linear structures such as portraits, resulting in undesired distortions in those regions. Recently, the learning solutions addressed this issue by forcing the perceptual semantics to be natural. In particular, Nie \textit{et al.} proposed the first deep rectangling pipeline \cite{nie2022deep}  with a coarse-to-fine mesh prediction architecture and conjoint constraints regarding boundary, distortion, and content. To train and evaluate it, the first rectangling benchmark (DIR-D) is also developed. Afterward, Liao \textit{et al.} proposed RecRecNet \cite{liao2023recrecnet} for rectangling rectified wide-angle images with TPS representation and DoF-based curriculum learning.

\subsection{Portrait Correction}
Wide-angle cameras are popular in smartphones as they can capture a wider field-of-view (FoV). However, severe wide-angle distortions are produced around the boundary regions, which could bend straight lines on buildings and distort faces \cite{cooper2012perceptual, fried2016perspective, beeler2010high}. To this end, several traditional algorithms \cite{tehrani2013undistorting, tehrani2016correcting} are proposed to remove such distortions in foreground objects minimally distorting the background, but this process demands assistance from a user interface. The more related works lie on \cite{shih2019distortion, lai2021correcting}, which strike a balance between foreground portraits and background straight lines through the combination of stereographic and perspective projections. Nevertheless, the camera parameters and portrait segmentations are required in advance. Besides, Zhang \textit{et al.} leveraged content-aware conformal mapping \cite{zhang2023wide} to optimize the polar mesh-based energies. But it also requires visual features (\textit{e.g.}, detected curves and salient maps) as extra input.%, and can only preserve the existing portrait instead of correcting the face distortions.

In contrast, the current learning solutions are free from these input limitations and head for simultaneous correction regarding lines and portraits. Accordingly, Tan \textit{et al.} \cite{tan2021practical} proposed a two-stage network and a portrait dataset with pairs of wide-angle images and corresponding correction flows, converting this task into estimating mapping flows. Later, to reduce the manual cost of preparing such a dataset, Zhu \textit{et al.} \cite{zhu2022semi} developed a semi-supervised transformer by introducing a surrogate segmentation task.

\vspace{0.2cm}
\noindent\textbf{\colorbox[rgb]{0.93,0.93,0.93}{Summary}}
Although these tasks have different goals, all of these are solved by estimating the desired warps. In light of this, we try to unify them in the same framework. Moreover, all solutions overlooked a basic observation that the output could be regarded as another input with less difficulty, \textit{e.g.}, an imperfect rotation correction result can be seen as an input with a smaller tilted angle. Based on this observation, we propose CoupledTPS to iteratively couple multiple TPS transformations, yielding better flexibility and performance.

	\section{Methodology}
\label{sec:Methodology}
In this section, we first justify the reason we chose TPS and its performance bottleneck. Then we describe the details of the CoupledTPS model and semi-supervised learning strategy. Finally, we briefly depict the extensions to rectangling and portrait correction tasks.

\begin{figure}[!t]
  \centering
  \includegraphics[width=.43\textwidth]{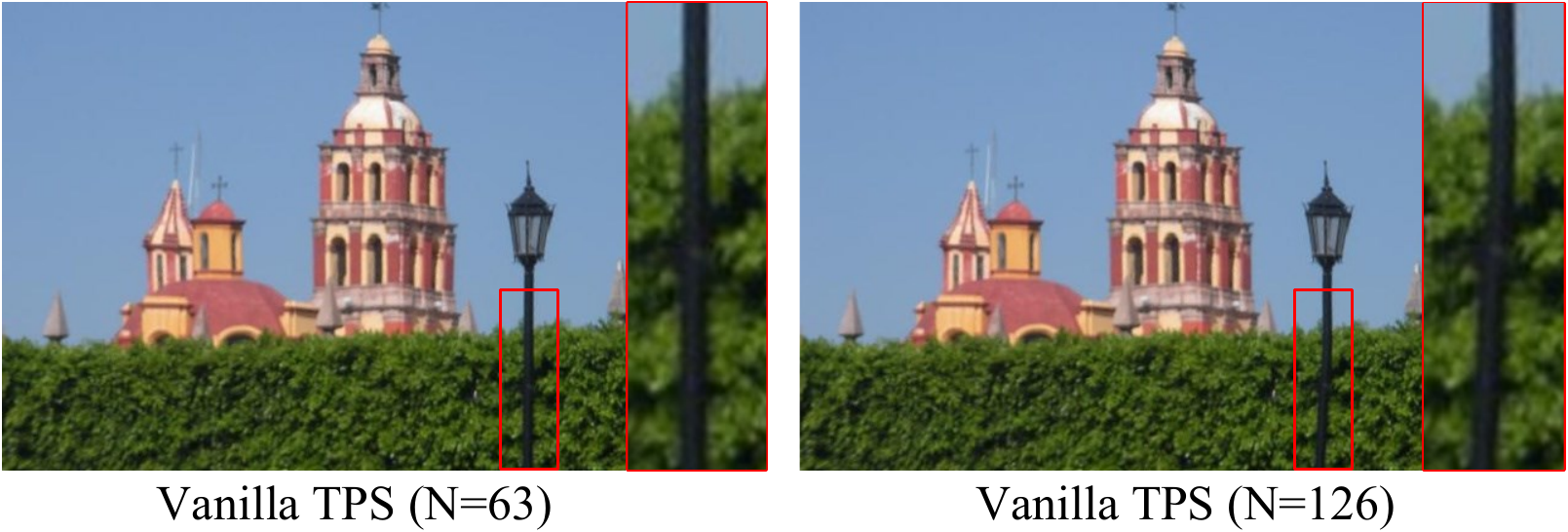}
  %\vspace{-20pt}
  \vspace{-0.2cm}
  \caption{The visualization of the performance bottleneck in vanilla TPS. With the increase of the control point number, the corrected results are prone to produce content distortions.}
  \label{fig:distortion}
  \vspace{-0.2cm}
\end{figure}

\begin{figure*}[!t]
  \centering
  \includegraphics[width=.97\textwidth]{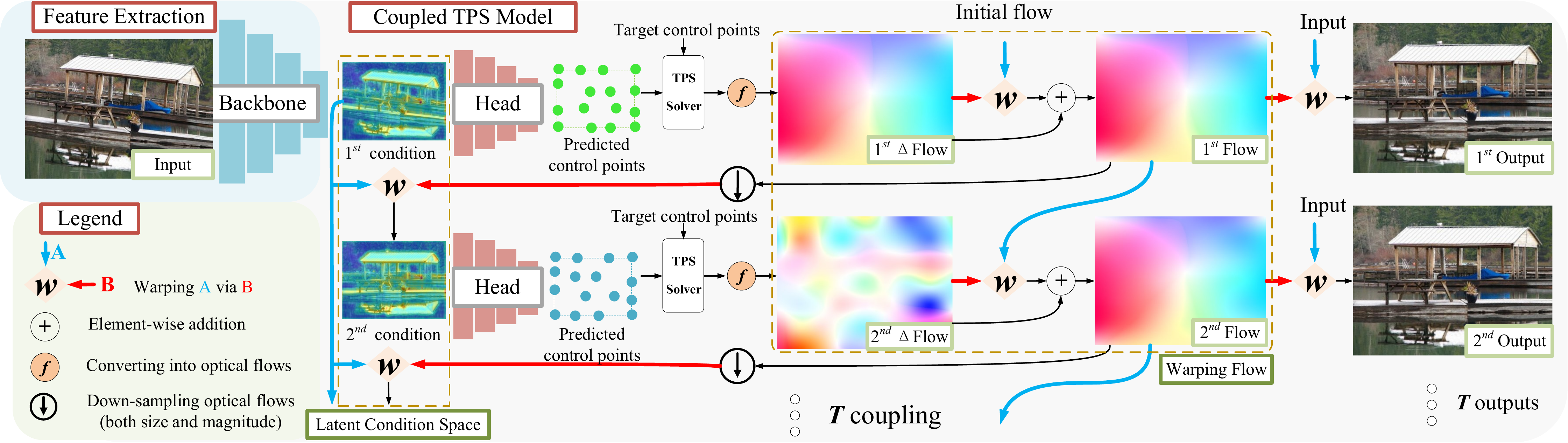}
  %\vspace{-20pt}
  \caption{The workflow of the proposed CoupledTPS. It first encodes an input image into the latent condition and then predicts the source control points from the latent condition. The predicted warp for each iteration is used to update the latent condition for the next iteration. The warping flow is leveraged to couple the currently predicted warp with the previously coupled warp and eliminate interpolation errors. The initial flow is set to 0.}
  \label{fig:network}
  \vspace{-0.4cm}
\end{figure*}

\begin{table}[!t]
  \centering
  \caption{The performance bottleneck of TPS. The performance encounters a bottleneck when the number of control points increases to 63.}
  \vspace{-0.2cm}
  \renewcommand{\arraystretch}{1.2}
  \begin{tabular}{cccc}
   \toprule
   & Control point number & PSNR ($\uparrow$) & SSIM ($\uparrow$) \\
   \cline{2-4}
 1 & 12 & 21.00 & 0.621  \\
 2 & 20 & 21.43 &  0.639 \\
 3 & 35 & 21.83 & 0.665 \\
 4 & 63 & \bfseries 22.04 & \bfseries 0.667 \\
 5 & 88 &  22.03 & 0.667 \\
 6 & 126 & 22.01 & 0.665 \\
      \bottomrule
   \end{tabular}
   \vspace{-0.1cm}
   \label{table:problem}
   \end{table}

\subsection{Problem Formulation}
{\noindent\textbf{\colorbox[rgb]{0.93,0.93,0.93}{Thin-Plate Spline}}
TPS is used to simulate 2D deformation specified by a set of source control points ($\bm{P} = [p_1, p_2, \cdots, p_N]^T$, $p_i\in \mathbb{R}^{2\times 1}$) and the corresponding target ones ($\bm{Q} = [q_1, q_2, \cdots, q_N]^T$, $q_i\in \mathbb{R}^{2\times 1}$ ). To approximate such a non-linear transformation with minimum non-rigid distortion, it is formulated as:
\begin{equation}
  \label{eq:tps1}
  \begin{matrix}
    \begin{aligned}
      \min \iint_{\mathbb{R}^{2}}&\left(\left(\frac{\partial^{2} \mathcal{T}}{\partial x^{2}}\right)^{2}\right.+2\left(\frac{\partial^{2} \mathcal{T}}{\partial x \partial y}\right)^{2}  +\left. \left(\frac{\partial^{2} \mathcal{T}}{\partial y^{2}}\right)^{2}\right) dx dy,\\
       &s.t. \quad \mathcal{T}(p_i) = q_i, i = 1,2,\cdots,N,
    \end{aligned}\\ 
  \end{matrix}
\end{equation}
where $\mathcal{T}(\cdot)$ represents the desired transformation. Then we can reach a spatial mapping function according to the derivation in \cite{bookstein1989principal} as follows:
\begin{equation}
  \mathcal{T}(p)=C+Mp+\sum_{i=1}^{N}w_iO(\parallel p-p_i\parallel_2 ),
  \label{eq:tps2}
\end{equation}
where $p$ is an arbitrary point on the source image and $O(r)=r^2logr^2$ is a radial basis function that indicates the impact of each control point on $p$. $C\in \mathbb{R}^{2\times 1}$, $M\in \mathbb{R}^{2\times 2}$, and $w_i\in \mathbb{R}^{2\times 1}$ are the transformation parameters.
%, which can be solved with $N$ pairs of control points and dimensional constraints \cite{kent1994link}.

To solve these parameters, we can formulate $N$ data constraints with $N$ pairs of control point correspondences ($i.e.$, $\bm{P}, \bm{Q} \in\mathbb{R}^{N\times 2}$) following Eq. \ref{eq:tps2}. Besides, the dimensional constraints \cite{bookstein1989principal},\cite{kent1994link} are leveraged to assist the solving process as follows:
\begin{equation}
  \sum_{i=1}^{N}w_i=0 \hspace{0.2cm} , \hspace{0.2cm} \sum_{i=1}^{N}p_iw_i^T=0.
  \label{eq:tps3}
\end{equation}
Then, we can rewrite all constraints (including data and dimensional constraints) as a form of matrix calculation and solve the parameters as:
\begin{equation}
    \begin{bmatrix} C^T\\ M^T \\W \end{bmatrix}=
    \begin{bmatrix} 1_{N\times1} &\bm{P}& K\\ 0_{1\times1} &0_{1\times2} &1_{1\times N} \\0_{2\times1} &0_{2\times2}& \bm{P}^T \end{bmatrix}^{-1}\begin{bmatrix} \bm{Q} \\0_{2\times2} \\0_{1\times2}\end{bmatrix},
    \label{eq:tps4}
 \end{equation}
where $W=[w_i,...,w_N]^T$. Each element $k_{ij}$ in $K\in \mathbb{R}^{N\times N}$ is determined by $O(\parallel p_i-p_j\parallel_2 )$.

%\subsubsection{TPS vs. Flow vs. Mesh}
\vspace{0.2cm}
\noindent\textbf{\colorbox[rgb]{0.93,0.93,0.93}{TPS vs. Flow vs. Mesh}}
%首先justify 为什么用TPS，它相比于mesh （更smooth）和 flow （更鲁棒）的优势。
As proved in \cite{nie2023deep}, the mesh is more robust than the flow in the task of rotation correction due to its sparsity. Similarly, if we place the control points on the grid vertices of the mesh, TPS can also be fulfilled with the sparsity and robustness of the mesh. However, in the mesh deformation, each point is only affected by the four nearest neighbor vertices, which inevitably leads to unsmooth transitions between adjacent grids. In contrast, TPS is free from this issue because each point is affected by all the control points as described in Eq. \ref{eq:tps2}. To this end, we choose TPS instead of the flow or mesh because of its advantages of robustness and smoothness.

To unify rotation correction, rectangling, and portrait correction in a uniform framework, we define the target control points fixedly distributed on the warped image and then predict the source control points on the input image.

% \subsubsection{Bottleneck}
\vspace{0.2cm}
\noindent\textbf{\colorbox[rgb]{0.93,0.93,0.93}{Bottleneck}}
As control points increase, TPS becomes increasingly flexible. However, its warping performance usually increases first and then encounters a bottleneck. Table \ref{table:problem} reveals this phenomenon, where we adopt a basic network consisting of a backbone and a head to predict the source control points for rotation correction. As this table shows, the performance is stuck when the control point number reaches 63. This bottleneck is because excessive control points are prone to produce some unexpected issues, e.g., content distortions. A qualitative example is depicted in Fig. \ref{fig:distortion}, where the lamp pole is bent due to excessive control points. In fact, more control points will produce more significant distortion. Here we only show the results of the control point amounts set to 63 and 126 for clarity.
% This might be caused by two reasons:
% \vspace{-0.2cm}
% \begin{itemize}
%     \item Some warping tasks themselves do not require extraordinarily flexible transformations. For example, in portrait correction, we only need to fine-tune the faces and bent lines in the local boundary regions.
%     \item Excessive control points might lead to unexpected issues, e.g., content distortion.
%  \end{itemize}

\subsection{Coupled Thin-Plate Spline Model}
To break the bottleneck as mentioned above, we propose CoupledTPS to simultaneously enhance the warping flexibility and performance. As illustrated in Fig. \ref{fig:network}, CoupledTPS has two significant differences from the basic TPS model: iterative search for new control points and warping flow for coupling.

\begin{figure}[!t]
  \centering
  \includegraphics[width=.47\textwidth]{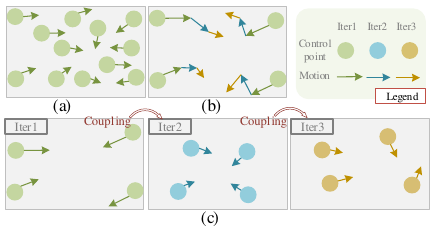}
  %\vspace{-20pt}
  \vspace{-0.2cm}
  \caption{Comparison of control point prediction strategies. (a) Predicting massive control points at once. (b) Predicting residual motions iteratively like RAFT \cite{teed2020raft}. (c) Predicting different control points iteratively (ours).}
  \label{fig:iter}
  \vspace{-0.2cm}
\end{figure}

% \subsubsection{Iteration}
\vspace{0.2cm}
\noindent\textbf{\colorbox[rgb]{0.93,0.93,0.93}{Iterative Search}}
In most cases, the outputs of the algorithms are not as perfect as we expect. For instance, the result of rotation correction might still contain tilt with a smaller magnitude. 
%the output of rectangling might leave some irregular boundaries, and the correction of wide-angle portraits might be slightly distorted. 
In light of this, we propose to treat the original input image as a hard sample and the rectified result as an easy sample. Then the easy sample can be seen as another input of the algorithm, yielding an iterative correction procedure.

%每次迭代预测新的控制点
In each iteration, the procedure is required to search for new source control points with shared network parameters, as demonstrated in Fig. \ref{fig:iter}c. Since the searched control points per iteration are different, the total transformation is viewed as a complicated coupled warp from multiple basic TPS transformations with limited control points. 
% In this way, the network only needs to predict limited control points, ensuring the robustness and effectiveness of each iteration. At the same time, the newly predicted control points increase the flexibility of the coupled transformation, which is different from predicting residual motions of the same control points at each iteration like RAFT \cite{teed2020raft}.
Note it differs from iteratively predicting residual motions of the same control points like RAFT \cite{teed2020raft} (shown in Fig. \ref{fig:iter}b).
Moreover, due to the limited number of control points at each iteration, content distortion (Fig. \ref{fig:distortion}) caused by extensive control points (Fig. \ref{fig:iter}a) is largely prevented but warping flexibility is boosted.

To accelerate the iterative search, we replace the high-resolution input image with the low-resolution latent condition. As shown in Fig. \ref{fig:network}, we leverage a backbone network to encode the input image into high-level semantic feature maps, which is termed the latent condition. Then a regression head is designed to predict the source control points from the latent condition, which formulates a TPS warp with predefined target control points. In the next iteration, the current latent condition is directly obtained by updating the previous condition rather than re-encoding the previous output image using the backbone. 

For the network, we adopt ResNet18 \cite{he2016deep} as our backbone, and the latent condition has the size of 1/16 of the input image resolution. The regression head is stacked by six convolutions, three max pooling, and three fully connected layers to predict the coordinates of $N$ source control points.

% \vspace{0.2cm}
% \noindent\textbf{\colorbox[rgb]{0.93,0.93,0.93}{Latent Condition Space}}
% If we directly use the output image of the previous iteration as the input image of the next iteration, the computational complexity of the entire procedure will increase linearly as the iterations increase, e.g., the complexity of three iterations is three times that of a single iteration. To reduce the computational cost and accelerate the inference, we define the latent condition space to replace the input image. As shown in Fig. \ref{fig:network}, we leverage a backbone network to encode the input image into high-level semantic feature maps, which we term the latent condition. Then a regression head is designed to predict the source control points from the latent condition, formulating a TPS warp with the predefined target control points. In the next iteration, the current latent condition is directly obtained by updating the previous condition rather than re-encoding the previous output image using the backbone. Each iteration shares the common parameters of the regression head.

% In particular, we adopt ResNet18 \cite{he2016deep} as our backbone and the latent condition has the size of 1/16 of the input image resolution. The regression head is stacked by six convolutions, three max pooling, and three fully connected layers to predict the coordinates of $N$ source control points.

\begin{figure*}[!t]
  \centering
  \includegraphics[width=.94\textwidth]{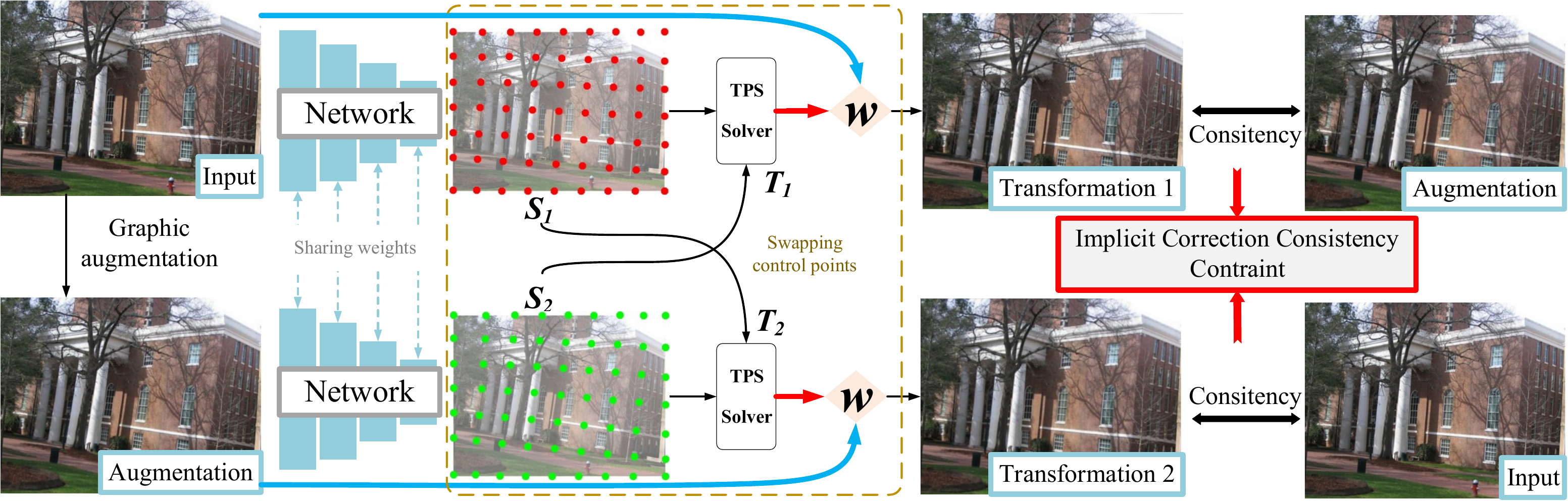}
  %\vspace{-20pt}
  \caption{Dual transformation for unlabeled data. We design the dual transformation between the unlabeled data and its graphic augmentation to establish the implicit correction consistency constraint.}
  \label{fig:mutual}
  \vspace{-0.4cm}
\end{figure*}

% \subsubsection{Warping Flow}
\vspace{0.2cm}
\noindent\textbf{\colorbox[rgb]{0.93,0.93,0.93}{Warping Flow}}
Due to the different source control points predicted from each iteration, each TPS transformation has different parameters. Denoting the predicted $i$-th TPS warp as $\mathcal{T}_i(\cdot)$, the inverse transformation from the target points to source points could be described as $\mathcal{T}^{-1}_i(\cdot)$. Supposing the original input image has a resolution of $H\times W$, we can define the grid $G$ of discrete pixel coordinates as follows: 
\begin{equation}
  G =
  \begin{Bmatrix}
    (x, y)|x = 1,2,\cdots,W; y = 1,2,\cdots,H
  \end{Bmatrix}.
  \label{eq:grid}
\end{equation}
Then $i$-th iteration output image $I_i$ is obtained by:
\begin{equation}
  I_i = I_{i-1}(\mathcal{T}^{-1}_i(G)),
  \label{eq:iter1}
\end{equation}
where $I_0$ represents the original input image at the first iteration. As indicated in Eq. \ref{eq:iter1}, to obtain the result of the $i$-th iteration, it requires $i$ image interpolation operations from $I_0$, $I_1$, $\cdots$ to $I_i$. Inevitably, this iterative procedure will produce interpolation blurs to different degrees according to the iteration number, degrading the image quality.

To eliminate these interpolation errors caused by multiple warps, we propose the warping flow to couple multiple TPS transformations into a whole. Denoting $\Delta F_i$ and $F_i$ are the converted flows from the current TPS transformation and the final warping flows in the $i$-th iteration, the iterative coupling process can be described as follows:
\begin{equation}
  \label{eq:iter2}
  \begin{matrix}
    \begin{aligned}
      & \Delta F_i = \mathcal{T}^{-1}_i(G) - G,\\
      & F_i = F_{i-1}(\mathcal{T}^{-1}_i(G)) + \Delta F_i,\\
      & I_i = I_0(G+F_i),
    \end{aligned}\\ 
  \end{matrix}
\end{equation}
where $F_0$ represents the initial flow that is set to 0. Note $\Delta F_{i}$ and $F_{i-1}$ cannot be directly added to get the final warping flow $F_{i}$ of the $i$-th iteration. Because these warping flows ($\Delta F_{i}$ and $F_{i-1}$) correspond to different latent conditions, we need to change the condition of $F_{i-1}$ from the last iteration into the current iteration first and then add them together. In this way, the $i$-th warp is coupled from the current TPS estimation and the previously coupled warp. As the number of iterations increases, more and more control points are coupled into a single warp (that's why we term CoupledTPS), while the interpolation number decreases from $i$ to 1, which effectively removes interpolation blurs.

\subsection{Dual Transformation for Correction Consistency}
In this section, we propose to establish an implicit correction consistency constraint to further boost the performance of labeled data from the easily available unlabeled data.

\vspace{0.2cm}
\noindent\textbf{\colorbox[rgb]{0.93,0.93,0.93}{Correction Consistency}}
For unlabeled data, it is an intuitive idea to contrast the correction results between the unlabeled input $I_u$ and its augmentation $I_u^{aug}$. Concretely, we can encourage the correction result of an unlabeled sample to approximate that of the augmentation to formulate an explicit correction consistency constraint. However, if we conduct the augmentation by injecting random illumination shifts on the image, the network tends to fall into a meaningless minimum. This is a situation in which the predicted source control points and predefined target control points completely coincide, yielding no change between the input and output. 

To this end, we propose to replace color augmentation with graphic augmentation. Particularly, we adopt content-aware rotation \cite{he2013content} to rotate an unlabeled tilted image with +/-3° randomly.
This non-rigid transformation can produce an augmented image with content tilted to a different degree from the original unlabeled input, while still maintaining regular boundaries with unnoticeable distortion.

\vspace{0.2cm}
\noindent\textbf{\colorbox[rgb]{0.93,0.93,0.93}{Dual Transformation}}
Despite that, this explicit correction consistency constraint can make the training process unstable, i.e., both sets of predicted source points are out of the image boundaries, yielding all-zero output images. Because both the correction results being contrasting are unfixed, which are changing according to network parameters.

To overcome this issue, we convert this explicit consistency constraint into an implicit one by dual transformation. Assuming that the correction results of $I_u$ and $I_u^{aug}$ are consistent, the two sets of predefined target control points in the two results correspond to identical pixels. Then the two sets of predicted source control points of $I_u$ and $I_u^{aug}$ should correspond to each other. Given this analysis, $I_u$ and $I_u^{aug}$ could be transformed mutually using their predicted source points. As shown in Fig. \ref{fig:mutual}, we take the source points $S_1$ of $I_u$ as the target points $T_2$ of $I_u^{aug}$ and take the source points $S_2$ of $I_u^{aug}$ as the target points $T_1$ of $I_u$. Subsequently, the implicit correction consistency constraint is formulated by dual transformation, which forces two transformed images close to the two original images. Moreover, this strategy would not result in unstable training because there is a clear and fixed transformation target for each sample ($I_u$ or $I_u^{aug}$), that is, each other.

\subsection{Semi-Supervised Learning}
The total objective function is composed of two terms according to the labeled and unlabeled data as follows:
\begin{equation}
  \mathcal{L}=\mathcal{L}_{l} + \mathcal{L}_{u}.
  \label{eq:total_loss}
\end{equation}

\vspace{0.2cm}
\noindent\textbf{\colorbox[rgb]{0.93,0.93,0.93}{Labeled Data}}
For an input image $I_0$ with the correction ground-truth $\widetilde{I_{0}}$, we define the labeled loss as follows:
\begin{equation}
  \mathcal{L}_{l} = \sum_{t=0}^{T-1}\gamma^t\mathcal{L}_{PE}(I_{t+1}, \widetilde{I_{0}}),
  \label{eq:label_loss}
\end{equation}
where $\gamma$ is set to 0.9 in our experiment. $\mathcal{L}_{PE}(\cdot,\cdot)$ is the perceptual loss \cite{johnson2016perceptual} in which we choose three different scales of VGG19 \cite{simonyan2014very} for perceptual measures.

\vspace{0.2cm}
\noindent\textbf{\colorbox[rgb]{0.93,0.93,0.93}{Unlabeled Data}}
For an unlabeled input $I_u$ and its graphic augmentation $I_u^{aug}$, we define the unlabeled loss as follows:
\begin{equation}
  \label{eq:unlabel_loss}
  \begin{matrix}
    \begin{aligned}
      \mathcal{L}_{u} = &\mathcal{L}_{PE}(I_u(\mathcal{T}_{S_2\to S_1}(G)), I_u^{aug}) \\[6pt]
      + &\mathcal{L}_{PE}(I_u^{aug}(\mathcal{T}_{S_1\to S_2}(G)), I_u),
    \end{aligned}\\ 
  \end{matrix}
\end{equation}
where $\mathcal{T}_{Src\to Tar}(\cdot)$ describes the TPS transformation from source control points to target control points.

\begin{table*}[!t]
    \centering
    \caption{The quantitative results of the proposed CoupledTPS and other rotation correction solutions on the DRC-D dataset \cite{nie2023deep}.}
    \vspace{-0.1cm}
    \label{table:rotation}
    \renewcommand{\arraystretch}{1.2}
    \begin{tabular}{cccccccc}
     \toprule
     &Method & Input & PSNR ($\uparrow$) &SSIM ($\uparrow$) &FID \cite{heusel2017gans} ($\downarrow$)&LPIPS \cite{zhang2018unreasonable} ($\downarrow$) & Time ($\downarrow$)\\
    %  \midrule
    \cline{2-8}
     1&Rotation & Tilted image \& angle & 11.57 &0.374 &34.40& 0.468 &-\\
     % 2& Rotation \& He et\ al.'s rectangling \cite{he2013rectangling} &  Yes &  17.63  & 0.488 & 15.30 & 0.345  \\
     % 3& Rotation \& Nie et\ al.'s rectangling \cite{nie2022deep}  &  Yes&  19.89  & 0.550 & 13.40 & 0.286  \\
     % 4& He et\ al.'s rotation \cite{he2013content } &  Yes &  21.69  & 0.646 & 8.51 & 0.212  \\
     2& Rotation \& He et al.'s rectangling \cite{he2013rectangling} &  Tilted image \& angle &  17.63  & 0.488 & 15.30 & 0.345&  -\\
     3& Rotation \& Nie et al.'s rectangling \cite{nie2022deep}  &  Tilted image \& angle&  19.89  & 0.550 & 13.40 & 0.286 &  65ms\\
     4& Content-aware rotation \cite{he2013content } &  Tilted image \& angle &  21.69  & 0.646 & 8.51 & 0.212&  - \\
     5& DRC \cite{nie2023deep}  & \bfseries Tilted image &  21.02  & 0.628 & \bfseries 7.12 & 0.205& 46ms \\
     % 6& CoupledTPS (ours)  & \bfseries Tilted image & \bfseries 22.39  & \bfseries 0.683 & 7.89 & \bfseries 0.196  \\
     %  6& CoupledTPS (ours) - revision & \bfseries Tilted image & \bfseries 22.29  & \bfseries 0.679 & 7.90 & \bfseries 0.197  \\
     6& CoupledTPS (ours) & \bfseries Tilted image & \bfseries 22.29  & \bfseries 0.679 & 7.90 & \bfseries 0.197& 38ms  \\

       \bottomrule
     \end{tabular}
    \vspace{-0.2cm}
     \end{table*}

\begin{figure*}[!t]
  \centering
  \includegraphics[width=.97\textwidth]{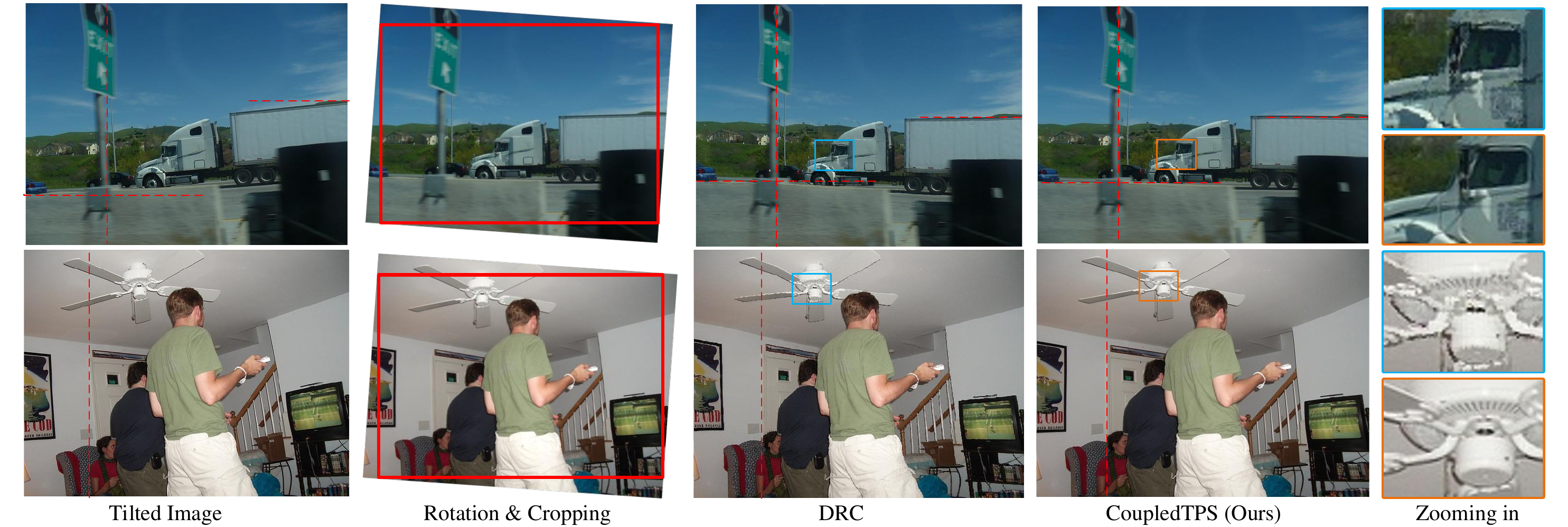}
  %\vspace{-20pt}
    \vspace{-0.2cm}
  \caption{Cross-domain comparisons on rotation correction. Both DRC \cite{nie2023deep} and CoupledTPS are trained on the DRC-D dataset and then evaluated on the MS-COCO dataset. The horizontal and vertical dashed lines are drawn to observe the tilt. } 
  \label{fig:rotation1}
\end{figure*}

\subsection{Extension to Other Tasks}
In addition to rotation correction, the proposed CoupledTPS can be easily extended to other tasks. Here, we describe the minor modifications to adapt to other fields.

% \vspace{0.2cm}
% \noindent\textbf{\colorbox[rgb]{0.93,0.93,0.93}{Rectangling}}
For rectangling, we take the concatenation of the stitched image and its mask as input, which is similar to \cite{nie2022deep}. For wide-angle portrait correction, we impose the optical flow constraint like \cite{zhu2022semi} because the portrait dataset \cite{tan2021practical} provides the correction flows as warp labels. Besides, considering that wide-angle distortions mainly occur around boundaries, we distribute control points more intensively on these regions instead of uniformly distributed over the entire image. 
%As for the semi-supervised learning scheme using dual transformation, it is not applied to these extended tasks, because there are no specific graphic augmentation technologies corresponding to rectangling and portrait correction (e.g., content-aware rotation \cite{he2013content} is used for rotation correction).

% \vspace{0.2cm}
% \noindent\textbf{\colorbox[rgb]{0.93,0.93,0.93}{Wide-Angle Portrait Correction}}

\section{Experiment}
\label{sec:Experiment}
In this section, we first introduce our training and inference details and then demonstrate extensive comparative experiments on three tasks. Subsequently, the ablation studies about CoupledTPS and semi-supervised learning are depicted to show their effectiveness. 

\subsection{Implement Detail}
For all the tasks, we train CoupledTPS with the labeled loss for about 120 epochs. Then, with the unlabeled data for rotation correction, we introduce the 
unlabeled loss with dual transformation, further fine-tuning CoupledTPS in the proposed semi-supervised scheme for about 60 epochs. The batch size is set to 4, and each batch only contains the labeled images or unlabeled images. The iteration number of the training stage is set to 4. All the training images are resized to a unified resolution of $384\times 512$. The control point numbers for rotation correction, rectangling, and portrait correction are set to 63, 63, and 82, respectively. The implementation is based on PyTorch and we use a single GPU of NVIDIA RTX 3090 Ti to finish both training and testing procedures.  

For the inference stage, the input images with arbitrary resolutions and arbitrary aspect ratios are supported. Concretely, we first resize them to $384\times 512$ to acquire the final warping flows. Then the warping flows are upsampled or downsampled (in both size and magnitude) to the original input resolution. The correction result can be obtained by warping the full-resolution input image using the resized deformation maps.

\begin{figure*}[!t]
  \centering
  \includegraphics[width=.97\textwidth]{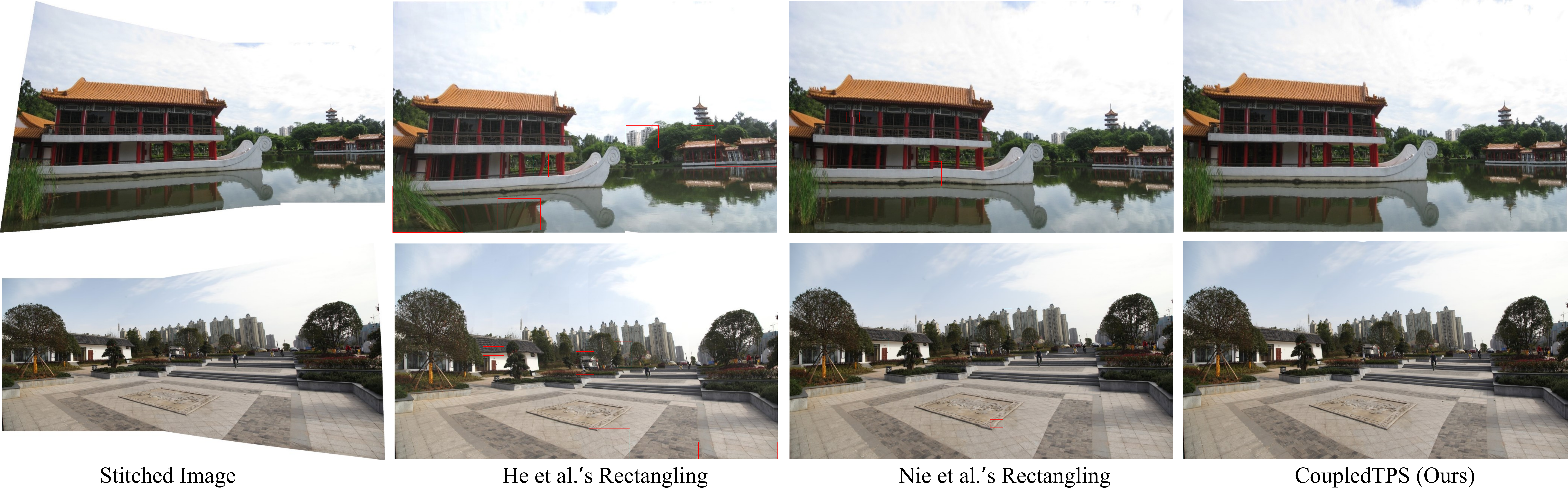}
  %\vspace{-20pt}
    \vspace{-0.2cm}
  \caption{Cross-domain comparisons on rectangling. Both Nie et al.'s rectangling \cite{nie2022deep} and CoupledTPS are trained on the DIR-D dataset and then evaluated on the classic image stitching dataset \cite{lin2016seagull, li2017parallax}. The red rectangles highlight the distorted regions. Zoom in for the best view. } 
  \label{fig:rectangling1}
\end{figure*}

\subsection{Comparative Experiment}
To demonstrate the effectiveness of CoupledTPS, we conduct extensive comparative experiments on the tasks of rotation correction, rectangling, and portrait correction.

\subsubsection{Rotation Correction}
For rotation correction, we conduct comparative experiments on the DRC-D dataset \cite{nie2023deep} and MS-COCO dataset \cite{lin2014microsoft}. The former is used to show the quantitative comparison, while the latter is adopted to depict the generalization capability. Specifically, the DRC-D dataset contains 5,537 samples for training and 665 samples for testing with the resolution of $384\times 512$. We train CoupledTPS on this dataset and further evaluate the generalization with real tilted images of different scenes and resolutions from MS-COCO. We compare CoupledTPS with rigid rotation, rotation+He et al.'s rectangling \cite{he2013rectangling}, rotation+Nie et al.'s rectangling \cite{nie2022deep}, content-aware rotation \cite{he2013content}, and DRC \cite{nie2023deep}, of which only DRC \cite{nie2022deep} and CoupledTPS do not require a specific tilted angle as input.

As illustrated in Table \ref{table:rotation}, the quantitative results on the DRC-D dataset \cite{nie2023deep} are evaluated from two aspects. On the one hand, PSNR and SSIM are selected to measure the objective image quality. On the other hand, we use FID \cite{heusel2017gans} and LPIPS \cite{zhang2018unreasonable} to show the perceptually semantic similarity. Compared with other solutions, the proposed CoupledTPS reaches better PSNR and SSIM by a large margin, which suggests our corrected results are closer to the ground truth. As for FID and LPIPS, CoupledTPS shows competitive performance with the existing SoTA solution.

Then we demonstrate the cross-domain comparisons in Fig. \ref{fig:rotation1}. In particular, we train these networks on the DRC-D dataset and then evaluate them qualitatively on the MS-COCO dataset with significantly different data distribution. As indicated by the auxiliary lines, the proposed solution can rectify the tilt slightly better. Moreover, our corrected results show sharper edges and more clear contents, especially in the cross-domain data. The generalization of CoupledTPS benefits from the sparsity of control points and iterative coupling mechanism. As for DRC, the flexible and dense optical flows are easy to be disturbed especially in unseen data, which makes the predicted flows among adjacent pixels not smooth and thus leads to warping blurs.

\begin{table}[!t]
  \centering
  \caption{The quantitative results of the proposed CoupledTPS and other rectangling solutions on the DIR-D dataset \cite{nie2022deep}.}
  \vspace{-0.2cm}
  \renewcommand{\arraystretch}{1.2}
  \scalebox{0.8}{
  \begin{tabular}{cccccc}
   \toprule
   & Method & PSNR ($\uparrow$) & SSIM ($\uparrow$) & FID \cite{heusel2017gans} ($\downarrow$) & Time ($\downarrow$)\\
   \cline{2-6}
 1 & Reference &  11.30&  0.325 & 44.47 & - \\
 % 2 & He et al.'s rectangling \cite{he2013rectangling}& 14.70 &  0.3775 &  38.19 \\
 % 3 & Nie et al.'s rectangling \cite{nie2022deep} &  21.28& 0.7141 & 21.77 \\
 2 & He et al.'s rectangling \cite{he2013rectangling}& 14.70 &  0.378 &  38.19&  -\\
 3 & Nie et al.'s rectangling \cite{nie2022deep} &  21.28& 0.714 & 21.77 &  65ms\\
 % 4 & CoupledTPS (ours)& \bfseries 22.09 & \bfseries 0.765 & \bfseries 20.11  \\
 %  4 & CoupledTPS (ours) - revision& \bfseries 22.09 & \bfseries 0.764 & \bfseries 20.02  \\
  4 & CoupledTPS (ours)& \bfseries 22.09 & \bfseries 0.764 & \bfseries 20.02 & 38ms\\
      \bottomrule
   \end{tabular}
   }
   \vspace{-0.1cm}
   \label{table:rectangling}
   \end{table}

% \begin{table}[!t]
%   \centering
%   \caption{The quantitative results of the proposed CoupledTPS and other rectangling solutions on the dataset of RecRecNet \cite{liao2023recrecnet}.}
%   \vspace{-0.2cm}
%   \renewcommand{\arraystretch}{1.2}
%   \scalebox{0.9}{
%   \begin{tabular}{cccccc}
%    \toprule
%    & Method & PSNR ($\uparrow$) & SSIM ($\uparrow$) & AP ($\uparrow$) & mIoU ($\uparrow$)\\
%    \cline{2-6}
%  1 & Cropping &  11.51&  0.191 & 21.8 & 18.6  \\
%  % 2 & He et al.'s rectangling \cite{he2013rectangling}& 14.70 &  0.3775 &  38.19 \\
%  % 3 & Nie et al.'s rectangling \cite{nie2022deep} &  21.28& 0.7141 & 21.77 \\
%  2 & Padding & 12.07 &  0.278 &  23.9 & 20.1 \\
%  3 & ROP \cite{liao2021towards} & 13.90 &  0.352 &  30.8 & 26.1 \\
%  %4 & Rectangling \cite{he2013rectangling} & 15.36 &  0.4211 &  34.7 & 30.2 \\
%  4 & RecRecNet \cite{liao2023recrecnet} & 18.68 &  0.545 &  41.3 & 37.8 \\
%  5 & CoupledTPS (old)& \bfseries 18.86 & \bfseries 0.570 & \bfseries  &  \\
%       \bottomrule
%    \end{tabular}
%    }
%    \vspace{-0.1cm}
%    \label{table:rectangling2}
%    \end{table}

\subsubsection{Rectangling}
Retangling aims to eliminate irregular boundaries and stretch them as a rectangle. Here, we focus on the application of image stitching \cite{nie2021unsupervised, nie2022learning, nie2020view}, which composites two or more images into an image with a wider FoV but introduces irregular boundaries. The boundaries of stitched images vary significantly according to the image contents and stitching algorithms. We compare our CoupledTPS with He et al's rectangling \cite{he2013rectangling} and Nie et al.'s rectangling \cite{nie2022deep} on the DIR-D \cite{nie2022deep} dataset. The DIR-D dataset has a wide range of irregular boundaries and scenes, which includes 5,839 samples for training and 519 samples for testing. Each image on this dataset has a resolution of $384\times 512$. The quantitative results are displayed in Table \ref{table:rectangling}, where ``Reference"  takes stitched images as rectangling results for reference. Compared with existing methods, CoupledTPS shows a clear superiority to them in all the metrics.

% For rectified fisheye images, the contents are shrunk towards the principal center according to the distortion parameters, exhibiting a visually narrow FoV and irregular shape. We conduct quantitative experiments with cropping, padding, ROP \cite{liao2021towards}, and RecRecNet \cite{liao2023recrecnet} on the dataset of RecRecNet. Among them, ``cropping" is the cropped result that directly discards the content around the deformed boundary, and ``padding" is the mirror padding to fill the blank region beyond the deformed boundary. The results are listed in Table \ref{table:rectangling2}, where AP and mIoU are leveraged to evaluate the detection and segmentation performance of the vision model \cite{he2017mask}. As we can observe, the evaluation results show that CoupledTPS outperforms the existing solutions not only in low-level visual reconstruction but also in high-level semantic recovery.

In addition, we conduct cross-domain comparisons on real image stitching datasets \cite{lin2016seagull, li2017parallax}. The rectangling results of stitched images are shown in Fig. \ref{fig:rectangling1}, where the input images are stitched by UDIS++ \cite{nie2023parallax}. As we can see, there are noticeable distortions in the results of He et al.'s rectangling due to its heavy reliance on line segment detection \cite{von2008lsd}. When line segments are missed or detected discontinuously, distortions usually appear. As for Nie et al.'s rectangling, we could find the discontinuous edges occurred at the junction of the adjacent grids. In contrast, CoupledTPS overcomes these issues through the extraction of robust semantic features and coupled smooth TPS transformations. 

\begin{table}[!t]
  \centering
  \caption{The quantitative results of the proposed CoupledTPS and other wide-angle rectification solutions on Tan et al.'s wide-angle portrait correction dataset \cite{tan2021practical}.}
  \vspace{-0.2cm}
  \renewcommand{\arraystretch}{1.2}
  \scalebox{0.85}{
  \begin{tabular}{ccccc}
   \toprule
   & Method & LineAcc \cite{tan2021practical} ($\uparrow$) & ShapeAcc \cite{tan2021practical} ($\uparrow$) & Time ($\downarrow$) \\
   \cline{2-5}
 1 & Shih et al. \cite{shih2019distortion} & 66.143 &  97.253 & -\\
 2 & Tan et al. \cite{tan2021practical} & 66.784 & 97.490 & -\\
 3 & Zhu et al. \cite{zhu2022semi} & \bfseries 66.825 &  97.491 & 687ms\\
 % 4 & CoupledTPS (old) & 66.730 & \bfseries 97.495 \\
 % 4 & CoupledTPS (new86) & 66.759  & \bfseries 97.498  \\
 % 4 & CoupledTPS (new96) & 66.808  & \bfseries 97.500  \\
   4 & CoupledTPS (ours)  & 66.808  & \bfseries 97.500 & 720ms \\
 %最后的指标仍在调试，还有机会提升，（new）记录的是目前调试的最优指标
 
      \bottomrule
   \end{tabular}
   }
   \vspace{-0.1cm}
   \label{table:portrait}
   \end{table}

\begin{figure*}[!t]
  \centering
  \includegraphics[width=.97\textwidth]{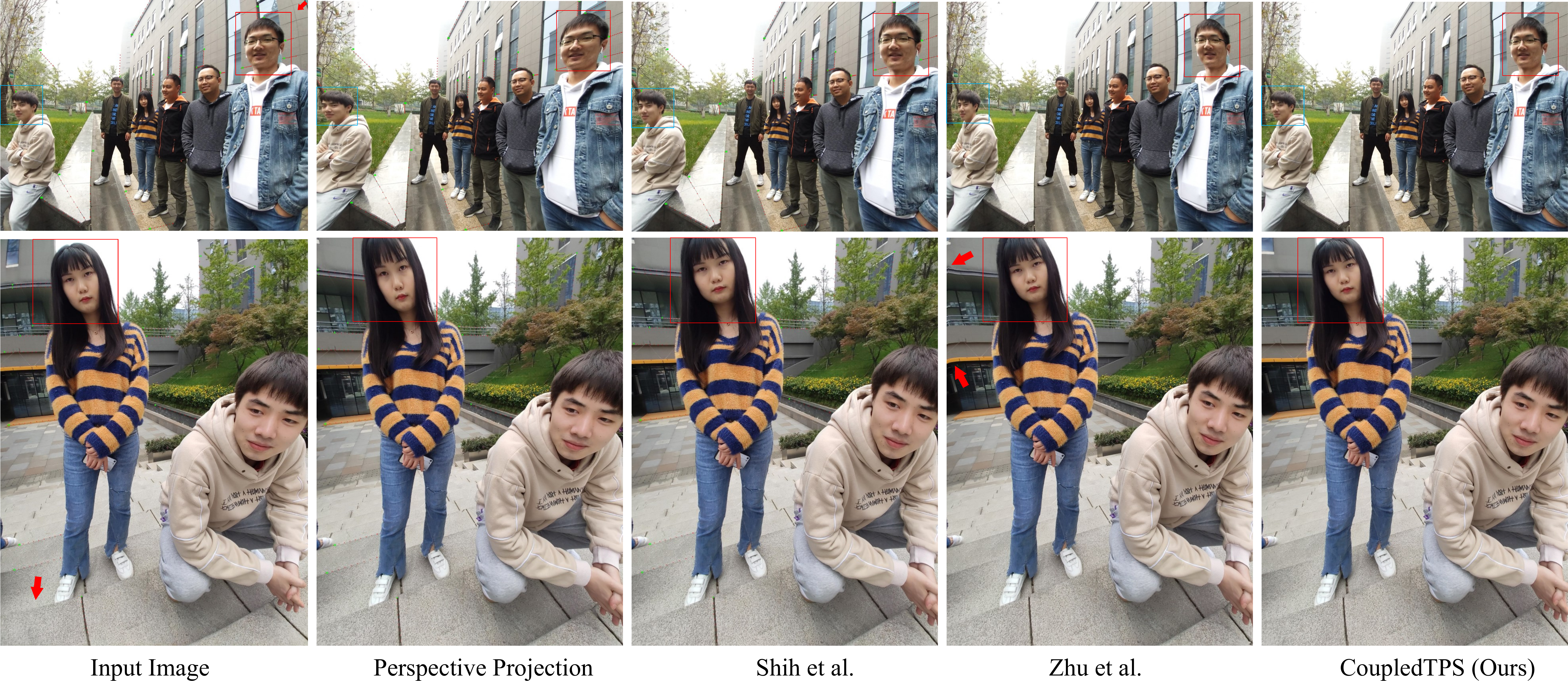}
  %\vspace{-20pt}
    \vspace{-0.2cm}
  \caption{Qualitative comparisons on portrait correction. These two examples are captured by Samsung A9S and Xiaomi 9, respectively. The arrows and rectangles highlight the distorted regions. Zoom in for the best view. } 
  \label{fig:portrait}
\end{figure*}

\begin{figure*}[t]
   \begin{center}
      \subfloat[Our results in rotation correction. Left: instances on the DRC-D dataset \cite{nie2023deep}. Right: instances on the unlabeled data from ImageNet \cite{deng2009imagenet}.]{\includegraphics[width=0.95\textwidth]{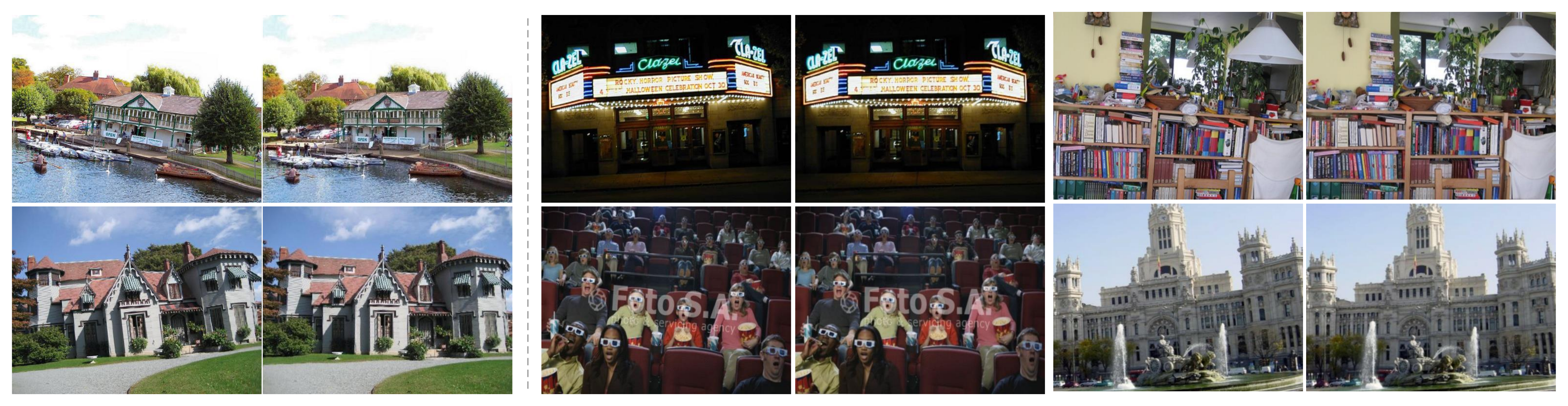}}
      \vspace{-0.1cm}
      \quad
      \subfloat[Our results in rectangling. Left: instances on the DIR-D dataset \cite{nie2022deep}. Right: instances on classic image stitching datasets \cite{lin2016seagull, li2017parallax, gao2011constructing}.]{\includegraphics[width=0.95\textwidth]{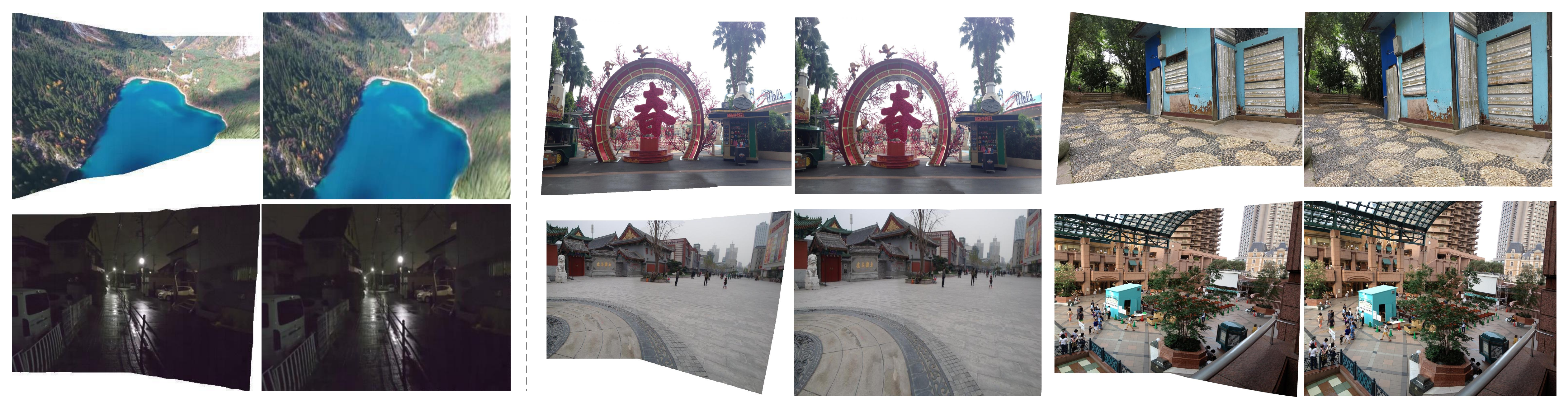}}
      \vspace{-0.1cm}
      \quad
      \subfloat[Our results in wide-angle portrait correction. Left: instances on Tan et al.'s dataset \cite{tan2021practical}. Right: instances on Zhu et al.'s dataset \cite{zhu2022semi}.]{\includegraphics[width=0.95\textwidth]{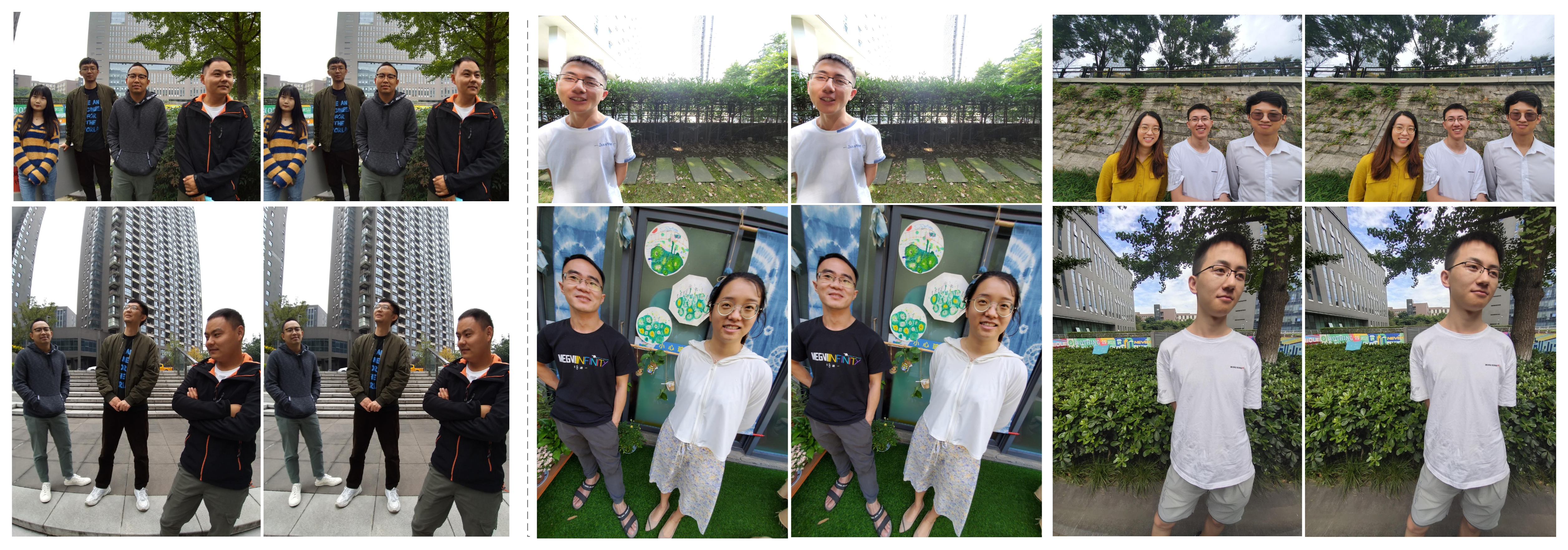}}
     \vspace{-0.1cm}
   \end{center}
   \caption{More results of CoupledTPS in different warping tasks. Each image pair consists of an input image and our warping result.}
   \vspace{-0.2cm}
\label{fig:more}
\end{figure*}

\subsubsection{Wide-Angle Portrait Correction}
For wide-angle portrait correction, we conduct the experiments on Tan et al's dataset \cite{tan2021practical}. The test set is captured from 5 types of smartphones including Samsung A9S, Samsung Note10, Vivo x23, Xiaomi 9, and Samsung A21. The wide-angle images cover different FoV, aspects, and resolutions (e.g., $3120\times 4160$, $3260\times 2448$, etc.). Following \cite{tan2021practical, zhu2022semi}, we adopt the same metrics (LineAcc and ShapeAcc) to evaluate the performance of our method. Concretely, LineAcc is used to evaluate the difference in curvature between the corrected lines and the corresponding marked lines in the perspective projection. ShapeAcc is used to assess the similarity between corrected portraits and that in the stereographic projection. 

We compare CoupledTPS with a traditional algorithm (Shih et al. \cite{shih2019distortion}) and two recent deep-learning solutions (Tan et al. \cite{tan2021practical} and Zhu et al. \cite{zhu2022semi}). The results are shown in Table \ref{table:portrait}, where our method reaches the best ShapeAcc among all the solutions. As for LineAcc, it cannot reflect the straightness of lines objectively because it implicitly requires that the corrected lines maintain the same slope as the lines in the perspective projection. If the corrected lines are straight but have an inconsistent slope with that in the perspective projection, it will produce a poor LineAcc score. Besides, some key bent lines are not marked (e.g., the bent lines in input images of Fig. \ref{fig:portrait}, as the arrows highlight), which further reduces the authority of this metric.

Then, we report the qualitative comparisons in Fig. \ref{fig:portrait}. We also show the perspective projected image of the input wide-angle image because Shih et al.'s solution takes the projected image as input. That's why the faces in the results of Shih et al. are relatively large, resulting in a slightly unharmonious proportion of the face and body. To obtain the perspective projected input image, the related camera parameters are required for Shih et al.'s solution. In contrast, Zhu et al.'s solution and CoupledTPS directly take the blind wide-angle image as input, free from the dependence of camera parameters. Moreover, there are some distortions in the boundaries of Zhu et al.'s results. Compared with them, CoupledTPS strikes the balance of faces and lines, producing more natural appearances.

\vspace{0.2cm}
\noindent\textbf{Inference Time} The comparison of inference time is also shown in Table \ref{table:rotation}\ref{table:rectangling}\ref{table:portrait}. To ensure experimental fairness, we only demonstrate the time of the methods that have released the official implementations. All the 
methods are tested on one RTX 3090 GPU. The inference time of our CoupledTPS may vary according to different resolutions in different tasks.

\vspace{0.2cm}
\noindent\textbf{More Result} In addition to the above comparative experiments, we demonstrate more results of CoupledTPS in Fig. \ref{fig:more} to show our universality and generalization. All the models are trained in the left datasets and further evaluated on cross-domain and cross-resolution other datasets.

\begin{figure}[!t]
  \centering
  \includegraphics[width=.47\textwidth]{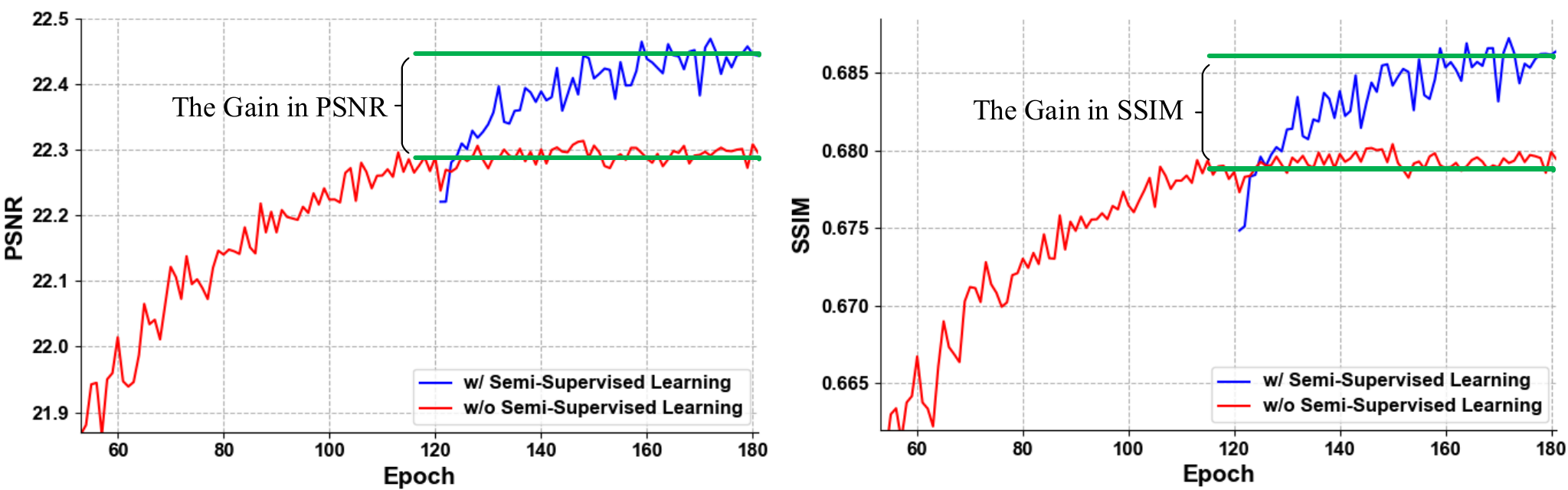}
  %\vspace{-20pt}
    \vspace{-0.2cm}
  \caption{The performance improvements of the semi-supervised learning on the labeled data.} 
  \label{fig:semi1}
\end{figure}

\begin{figure}[!t]
  \centering
  \includegraphics[width=.47\textwidth]{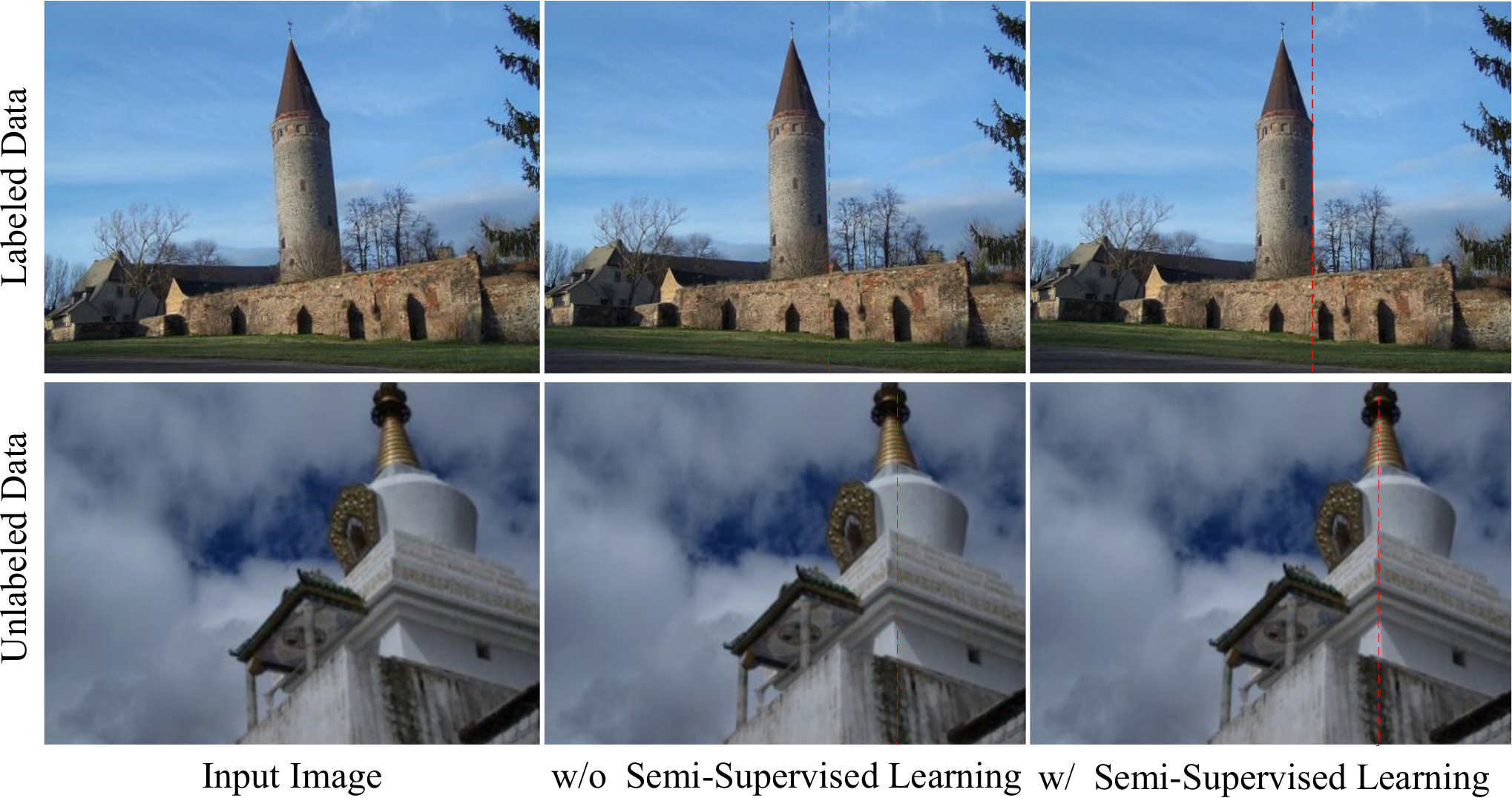}
  %\vspace{-20pt}
    \vspace{-0.2cm}
  \caption{The qualitative comparison between the corrected results with and without semi-supervised learning. The vertical dashed lines are drawn to observe the tilt.} 
  \label{fig:semi2}
\end{figure}

\begin{table}[!t]
  \centering
  \caption{ The performance improvements of the semi-supervised learning on the unlabeled data. The semi-supervised learning is abbreviated as SSL.}
  \vspace{-0.2cm}
  \renewcommand{\arraystretch}{1.2}
  \begin{tabular}{ccc}
   \toprule
    w/ SSL &  w/o SSL & No preference \\
   \hline
  14.4\% & 6.0\%  & 79.6\% \\
      \bottomrule
   \end{tabular}
   \vspace{-0.3cm}
   \label{table:semi1}
   \end{table}

\begin{table}[!t]
  \centering
  \caption{ The impact of the number of unlabeled images. The models are trained on the DRC-D dataset and collected unlabeled data. Then we quantitatively evaluate them on the DRC-D dataset.}
  \vspace{-0.2cm}
  \renewcommand{\arraystretch}{1.2}
  \begin{tabular}{cccc}
   \toprule
   & Number & PSNR ($\uparrow$) & SSIM ($\uparrow$) \\
   \cline{2-4}
 1 & 0  & 22.29 &  0.6786 \\
 2 & 500 & 22.31 & 0.6798 \\
 3 & 1000 & 22.33 & 0.6808 \\
 4 & 2000 & 22.40 & 0.6833 \\
 5 & 3000 & 22.42  & 0.6843 \\
 6 & 4000 & 22.44 & 0.6858 \\
 7 & 5000 & 22.44 & 0.6859 \\
 8 & 6760 & \bfseries 22.45 &\bfseries 0.6862 \\
      \bottomrule
   \end{tabular}
   \vspace{-0.1cm}
   \label{table:semi2}
   \end{table}

\subsection{Benefit of Semi-Supervised Learning}
Compared with labeled data, unlabeled data is easy to obtain since manual annotations are not required. To validate the effectiveness of the proposed semi-supervised learning, we collect about 7,516 images with different degrees of tilts from the ImageNet \cite{deng2009imagenet} dataset as the unlabeled dataset for rotation correction. Particularly, we take 6,760 images as the unlabeled training set, and the remaining are used to validate the performance improvements in the domain of unlabeled data.

We first demonstrate the performance improvement in the domain of original labeled data. The results are illustrated in Fig. \ref{fig:semi1}, where we draw quantitative indicators in the labeled testing set as the number of training epochs increases. With the proposed dual transformations for unlabeled data, the performance initially drops, then rises rapidly, and finally significantly outperforms that trained in the fully-supervised scheme. Compared with the fully supervised mode, our semi-supervised strategy increases the PSNR/SSIM by about 0.16/0.008.

Next, we study the improvement in the domain of unlabeled data. Since unlabeled data lacks a corrected ground truth, it is challenging to evaluate it quantitatively. As an alternative, we conduct a user study to compare the results before and after semi-supervised learning. In particular, we illustrate the corrected results before and after semi-supervised learning with a random order on a common screen and require the user to answer which image is more natural and perceptually horizontal. We invite 20 participants, including 10 researchers/students with computer vision backgrounds and 10 volunteers outside this community. For every participant, we randomly select a subset of 200 samples from the unlabeled testing set. The results are shown in Table \ref{table:semi1}, where the results after semi-supervised learning are much more preferred. Overall, the proposed dual transformations for semi-supervised learning not only improve the performance on the labeled dataset but also serve as an unsupervised domain adaption method to the target unlabeled dataset.

Besides that, we illustrate the results with and without semi-supervised learning in Fig. \ref{fig:semi2}. Having applied the dual transformation for semi-supervised learning, the tilt is better rectified.

Subsequently, we explore the impact of the number of unlabeled data by randomly sampling a certain amount of unlabeled images from the total unlabeled training set. The results are illustrated in Table \ref{table:semi2}, where the performance increases with the increase in the amount of unlabeled data in a specific range. 
%When the quantity exceeds 5$k$ (note the quantity of labeled training data is 5,537), the improvement becomes not noticeable.

\begin{table}[!t]
  \centering
  \caption{ Ablation studies of CoupledTPS. We validate the effectiveness of iterative search and warping flow on the DRC-D dataset.}
  \vspace{-0.2cm}
  \renewcommand{\arraystretch}{1.2}
  \begin{tabular}{ccccc}
   \toprule
   & Component & PSNR ($\uparrow$) & SSIM ($\uparrow$) & Time ($\downarrow$) \\
   \cline{2-5}
 1 & Inference iter 1 & 22.04  & 0.668 & 27ms\\
 2 & Inference iter 2 & 22.20 & 0.675 & 33ms\\
 3 & Inference iter 3 & 22.29 & 0.679  & 38ms\\
 4 & Inference iter 4 & 22.28 & 0.678 & 45ms\\
 %\cline{2-5}
 %5 & w/o Latent condition &  &   & 43ms\\
 \cline{2-5}
 6 & w/o Warping flow & 20.75 & 0.584 & 38ms\\
  \cline{2-5}
 7 & Vanilla TPS & 22.04  & 0.667 & 27ms\\
 8 & CoupledTPS  & 22.29 & 0.679 & 38ms\\
 9 & CoupledTPS (w/ SSL)  & 22.45 & 0.686 & 38ms\\
      \bottomrule
   \end{tabular}
   \vspace{-0.1cm}
   \label{table:ablation}
   \end{table}

\begin{figure}[!t]
  \centering
  \includegraphics[width=.47\textwidth]{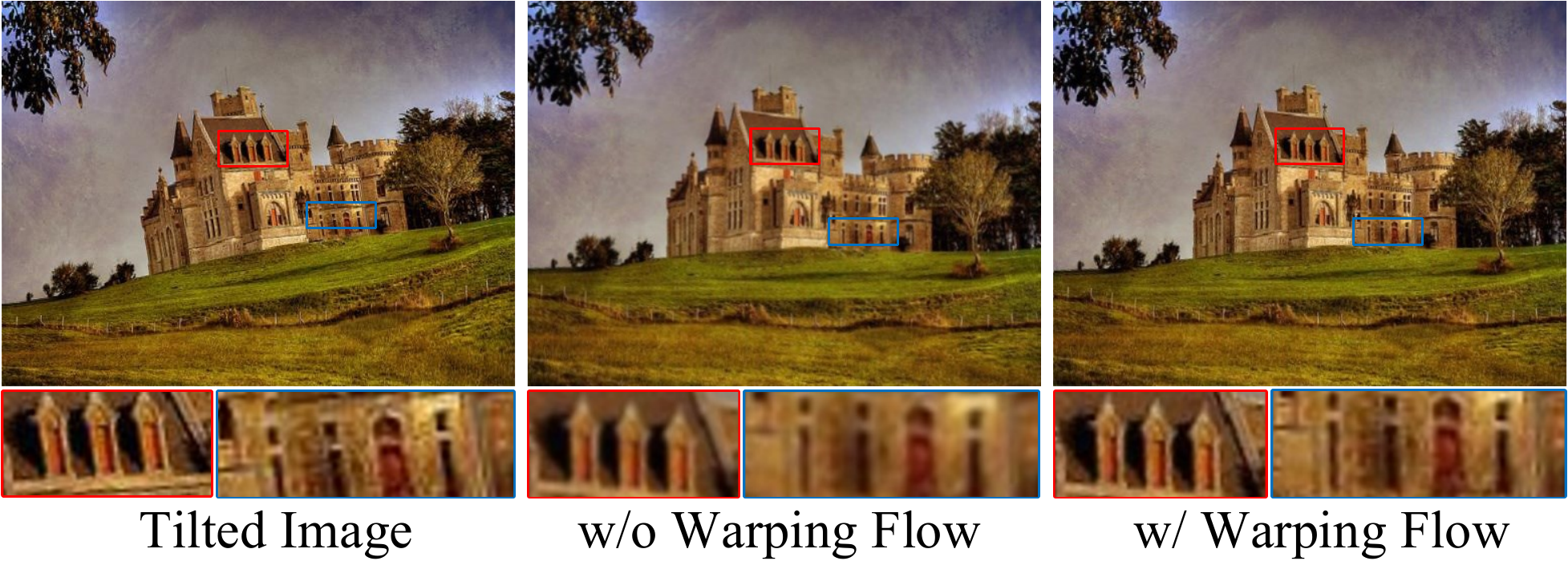}
  %\vspace{-20pt}
    \vspace{-0.2cm}
  \caption{Ablation study on warping flow. With warping flow, the corrected image shows sharper contours and clearer texture.
  %multiple basic TPS transformations are coupled as a whole, reducing the number of interpolation operations from $T$ to 1. 
  %yielding sharp contours and clear texture in the corrected image. 
  } 
  \label{fig:warpflow}
\end{figure}   

\subsection{Ablation Study}
Here, we discuss the effects of each module in CoupledTPS. The ablation experiments are conducted on the DRC-D dataset with a single GPU of RTX 3090Ti.

\vspace{0.2cm}
\noindent\textbf{Iteration Search.}
In each iteration, the proposed network is expected to search for new control points according to the current latent condition. As the iteration increases, the corrected result increasingly approximates the corrected ground truth. At this time, the newly predicted control points tend to overlap with the control points predicted in the previous iteration and thus the performance gains of each iteration will gradually decrease. Table \ref{table:ablation} reveals this phenomenon, where the $2^{nd}$ and $3^{rd}$ iterations bring the improvements of 0.16 and 0.09 in PSNR, respectively. When CoupledTPS is trained with $T$ set to 4, we get the best performance with the number of inference iterations set to 3. Moreover, with each additional iteration, the inference time only increases by $5\sim7$ms instead of the elapsed time of a complete iteration ($\sim27$ms), which proves the effectiveness of the latent condition space.

% \vspace{0.2cm}
% \noindent\textbf{Latent Condition.}
% The latent condition space is designed to reduce the computational complexity of multiple iterations. By directly updating the latent condition, we avoid re-encoding the result of the previous iteration, accelerating the inference speed. From the inference time shown in Table \ref{table:ablation}, we can find that a single iteration takes about 27ms. In contrast, each additional iteration (beginning with the latent condition) only increases the inference time by $5\sim7$ms instead of the whole time of another iteration ($\sim27$ms), which proves the effectiveness of the latent condition space.
 
\vspace{0.2cm}
\noindent\textbf{Warping Flow.}
The warping flow is designed to couple multiple TPS transformations into a whole and eliminate the interpolation blurs caused by multiple warps. To show its effectiveness, we ablate this module in CoupledTPS. Specifically,  we replace the coupled interpolation (Eq. \ref{eq:iter2}) with iterative interpolation (Eq. \ref{eq:iter1}). The quantitative results are shown in Table \ref{table:ablation}, where PSNR and SSIM decreased significantly without this module. In contrast, the inference time remains unchanged, which indicates the coupled interpolation process implemented by the warping flow does not bring additional computational complexity. A qualitative example is depicted in Fig. \ref{fig:warpflow}, where the interpolation number is reduced from 3 to 1 with the warping flow. The coupled interpolated result shows a noticeable superiority in sharp contours and clear texture.

\begin{table}[htbp]\centering
\setlength{\tabcolsep}{10pt}
% \tiny
\caption{ \centering Quantitative comparisons of different backbones on the DRC-D dataset.
}
\label{backbone}

% \resizebox{20cm}{5cm}
% \scalebox{1.4}
{
\begin{tabular}{c c| c c c c }
\hline
  & Backbone & PSNR & SSIM & Params & Time  \\
\hline
1 & ResNet18  & 22.29 & 0.679 & 116MB& 38ms \\
2 & ResNet34  &22.17 &0.671 &137MB & 40ms\\
3 & ResNet50  &22.27 & 0.678 &152MB & 41ms \\
4 & ResNet101  &22.24 &0.676 &225MB & 51ms \\
\hline     \hline     
5 & ViT-B/32  & 22.28& 0.677& 452MB& 36ms\\
6 & ViT-L/32  &22.41 & 0.685& 690MB& 46ms\\
7 & Swin-S & 22.47& 0.686& 236MB & 56ms \\
8 & Swin-B & 22.41&0.683 &338MB & 61ms\\
\hline 
\end{tabular}
}
\end{table}

\vspace{0.2cm}
\noindent\textbf{Backbone.}
We also conduct experiments to study the impact of diverse backbones on our performance. The results are illustrated in Table \ref{backbone}. In the series of ResNet \cite{he2016deep}, different convolutional layers (from 18 to 101) achieve similar performance, which implies the number of convolutional layers has a limited impact on performance. Besides, we also validate our CoupledTPS on Transformer-based backbones \cite{vaswani2017attention} (\textit{i.e.}, ViT \cite{dosovitskiy2020image} and Swin Transformer \cite{liu2021swin}). It shows that our CoupledTPS can produce better results on Transformer-based backbones. Interestingly, in ViT, the larger network (ViT-L/32) can produce better performance, while it is exactly the opposite in Swin Transformer. Considering the increase in computational complexity brought by transformers, we still chose ResNet18 as our backbone in this work. 
%We will further explore more types and variations of backbones in future work.

\vspace{0.1cm}
Note that all models shown in Table \ref{table:ablation} have identical sizes of parameters, which means the proposed CoupledTPS and semi-supervised learning scheme do not introduce additional parameters.

% \vspace{0.2cm}
% \noindent\textbf{Vanilla TPS vs. CoupledTPS.}
%  Compared with vanilla TPS, CoupledTPS gains the improvement of 0.25/0.012 in PSNR/SSIM at the cost of 11ms inference time. Note that all models shown in Table \ref{table:ablation} have identical parameter sizes of 116MB, which means the proposed CoupledTPS and semi-supervised learning will not increase the new parameters.

% \begin{figure}[!t]
%   \centering
%   \includegraphics[width=.4\textwidth]{figures/failure.pdf}
%   %\vspace{-20pt}
%     \vspace{-0.2cm}
%   \caption{Limitation of large tilt.} 
%   \label{fig:failure}
%   \vspace{-0.2cm}
% \end{figure}   

\subsection{Limitation}
The proposed dual transformation for semi-supervised learning requires a content-aware graphic argumentation approach. For rotation correction, we leverage the content-aware rotation algorithm \cite{he2013content} to generate an augmented image that has a different tilt, natural content, and regular boundary. As for the extended tasks (rectangling and wide-angle portrait correction), dual transformation is not applied because there are no specific graphic augmentation technologies corresponding to rectangling and portrait correction.
	\section{Conclusion}
\label{sec:Conclusion}
In this paper, we propose a coupled thin-plate spline model, termed CoupledTPS, to break the performance bottleneck of basic TPS transformation in single-image-based warping tasks. To this end, we design the iterative search to predict new control points from the latent condition, and the warping flow to couple multiple TPS transformations without interpolation blurs. 
Besides, to reduce laborious annotation costs and further improve the warping quality, a semi-supervised learning strategy is proposed to exploit the unlabeled data. Particularly, we establish an implicit correction consistency constraint through dual transformation between an unlabeled image and its graphic argumentation.
Extensive experiments show that CoupledTPS outperforms existing SoTA solutions in multiple warping tasks, such as rotation correction, rectangling, and wide-angle portrait correction. 
We also collect massive unlabeled data to validate the effectiveness of the proposed semi-supervised learning. The results show that it can boost the warping performance in both domains of labeled and unlabeled data in the rotation correction task. 

    \bibliographystyle{IEEEtran}
	\bibliography{bib}
 
	% \ifCLASSOPTIONcaptionsoff
	% \newpage
	% \fi
	
% \vspace{-1.8cm}
\begin{IEEEbiography}[{\includegraphics[width=1in,height=1.25in,clip,keepaspectratio]{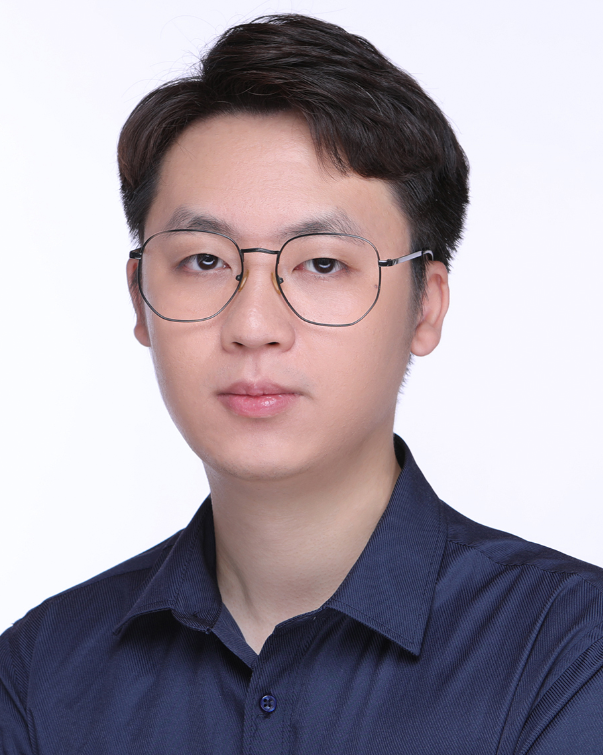}}]{Lang Nie} is currently pursuing the Ph.D.  degree in signal and information processing from the Institute of Information Science, Beijing Jiaotong University, Beijing, China. Prior to that, he received the B.S degree from Beijing Jiaotong University in 2019.
His current research interests include image and video processing, 3-D vision, and multi-view geometry.
\end{IEEEbiography}

\begin{IEEEbiography}[{\includegraphics[width=1in,height=1.25in,clip,keepaspectratio]{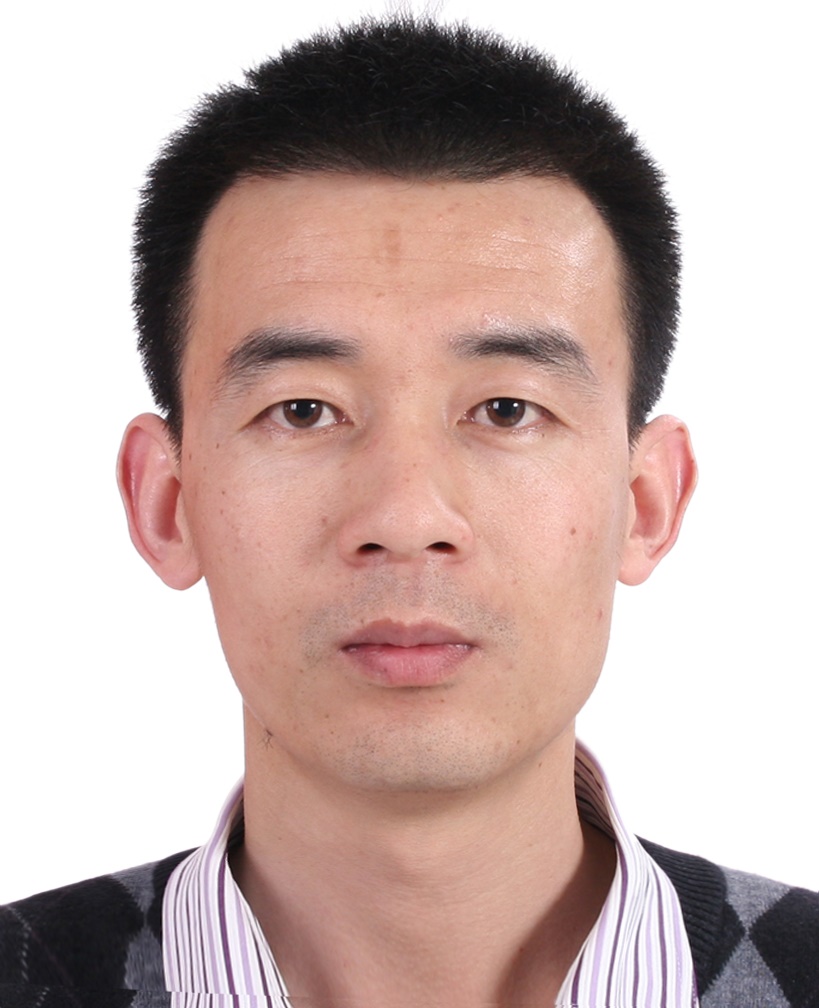}}]{Chunyu Lin}(Member, IEEE) is a Professor at Beijing Jiaotong University. He received the Doctor degree from Beijing Jiaotong University (BJTU), Beijing, China, in 2011. From 2009 to 2010, he was a Visiting Researcher at the ICT Group, Delft University of Technology, Netherlands. From 2011 to 2012, he was a Post-Doctoral Researcher with the Multimedia Laboratory, Gent University, Belgium. His research interests include multi-view geometry, camera calibration, and virtual reality video processing.
\end{IEEEbiography}

\begin{IEEEbiography}[{\includegraphics[width=1in,height=1.25in,clip,keepaspectratio]{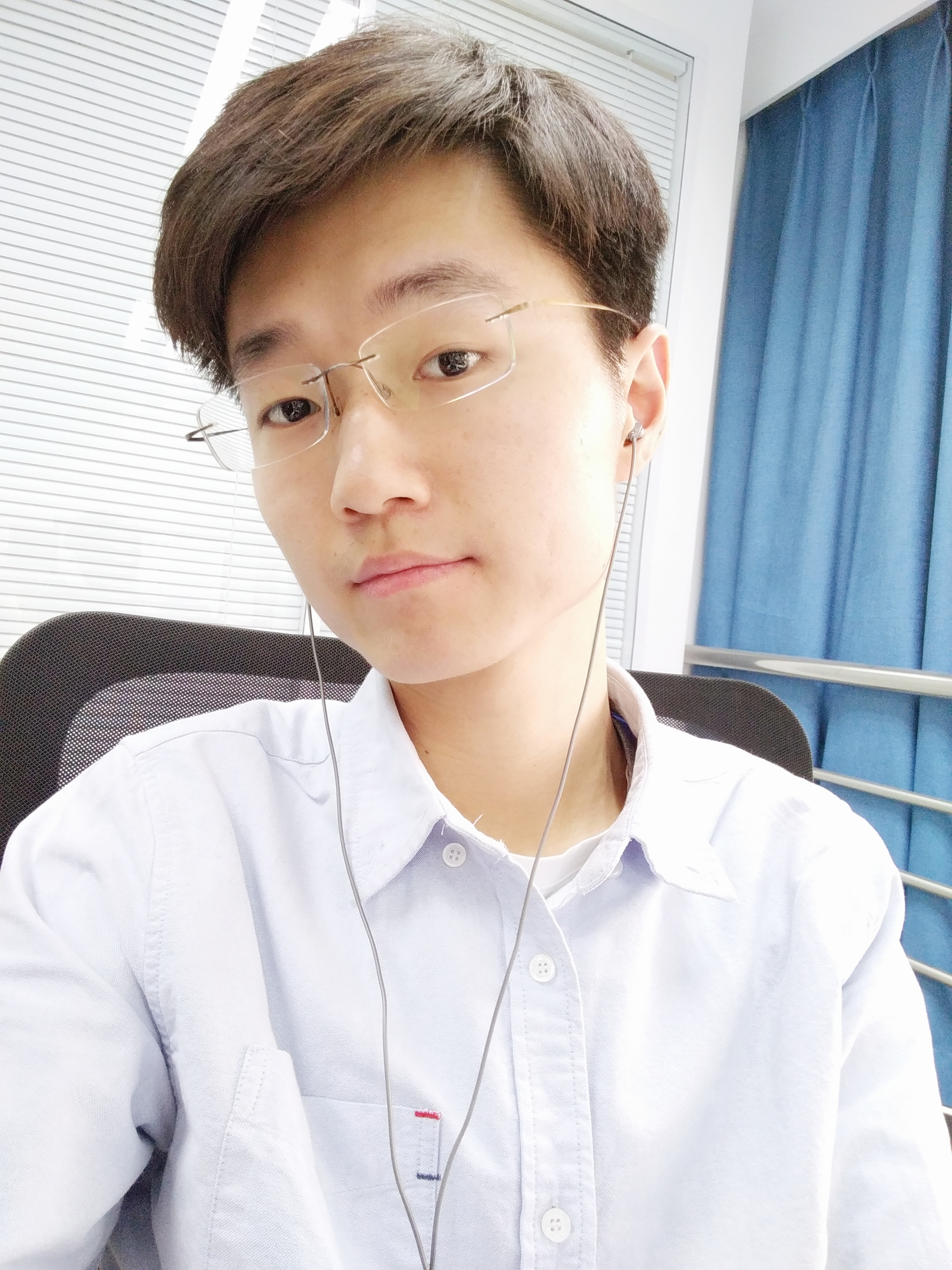}}]{Kang Liao} is currently a Research Fellow at Nanyang Technological University (NTU), Singapore. Before joining NTU, he received his Ph.D. degree from Beijing Jiaotong University in 2023. From 2021 to 2022, he was a Visiting Researcher at Max Planck Institute for Informatics in Germany. His current research interests include camera calibration, 3D vision, and panoramic vision.
\end{IEEEbiography}
               
\begin{IEEEbiography}[{\includegraphics[width=1in,height=1.25in,clip,keepaspectratio]{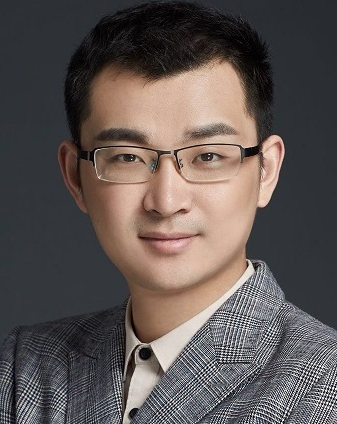}}]{Shuaicheng Liu} (Member, IEEE) is a Professor with the Institute of Image Processing, School of Information and Communication Engineering, University of Electronic Science and Technology of China (UESTC) since 2014. His research interests include computer vision and computer graphics. He received the Ph.D. and M.S. degrees from the National University of Singapore, Singapore, in 2014 and 2010, respectively. Before that, he received the B.E. degree from Sichuan University, Chengdu, China, in 2008.
\end{IEEEbiography}

\begin{IEEEbiography}[{\includegraphics[width=1in,height=1.25in,clip,keepaspectratio]{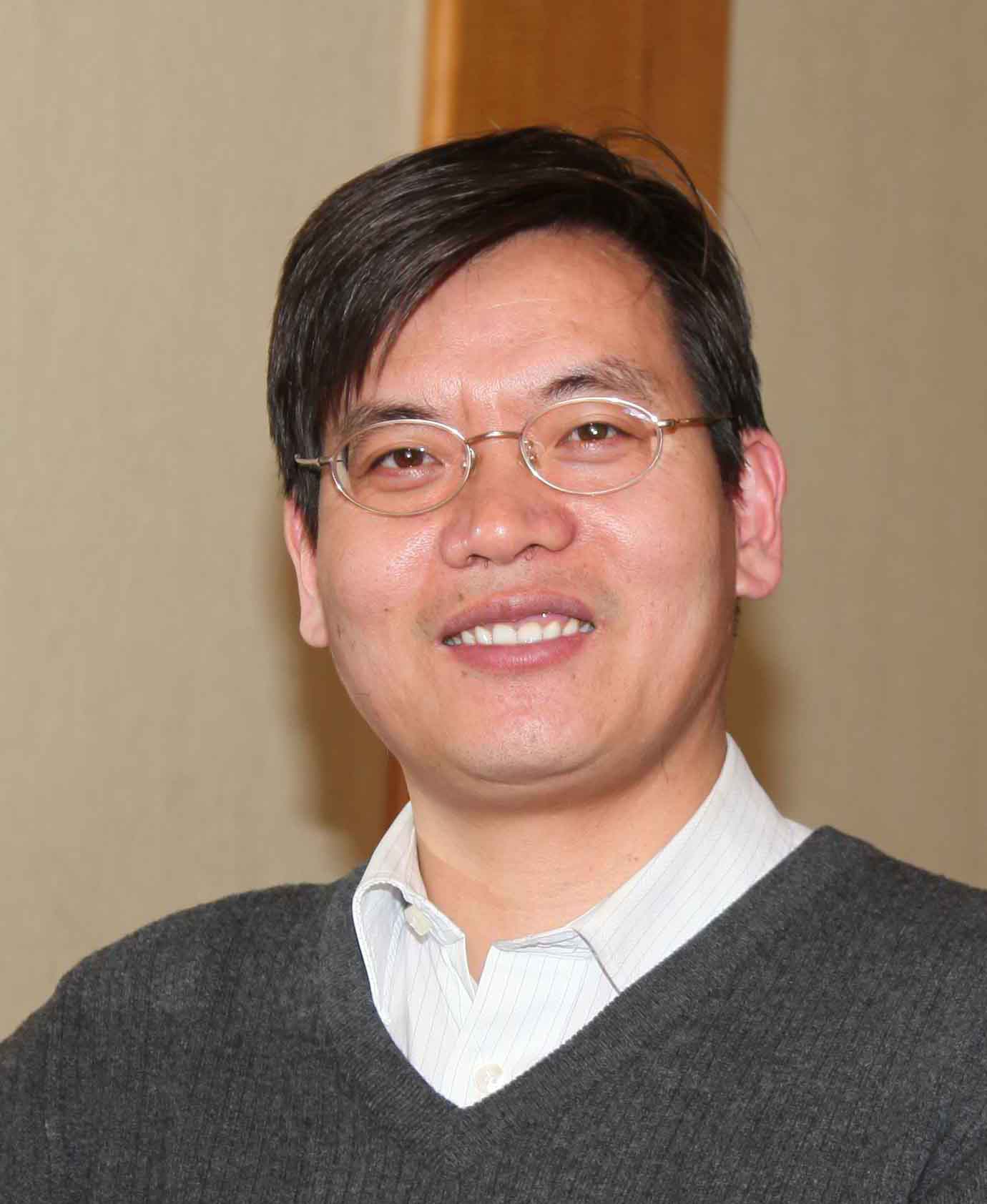}}]{Yao Zhao} (Fellow, IEEE) is currently the Director of the Institute of Information Science, Beijing Jiaotong University. His current research interests include image/video coding and video analysis and understanding. He was named a Distinguished Young Scholar by the National Science Foundation of China in 2010 and was elected as a Chang Jiang Scholar of Ministry of Education of China in 2013. Before that, he received the B.S. degree from Fuzhou University, Fuzhou, China, in 1989, and the M.E. degree from Southeast University, Nanjing, China, in 1992, both from the Radio Engineering Department, and the Ph.D. degree from the Institute of Information Science, Beijing Jiaotong University (BJTU), Beijing, China, in 1996. 
\end{IEEEbiography}

\end{document}